\documentclass[jounal]{acmsiggraph}

\usepackage{booktabs} % For formal tables

\usepackage[ruled]{algorithm2e} % For algorithms
\usepackage{amsmath}
\usepackage{amssymb}
\usepackage{mathrsfs}
\usepackage[british,UKenglish,USenglish,english,american]{babel}
\usepackage{booktabs}
\usepackage{color}
\usepackage{subfigure}
\usepackage{array}

\newcommand{\Tref}[1]{Table~\ref{#1}}
\newcommand{\Eref}[1]{Equation~(\ref{#1})}

\newcommand{\Sref}[1]{Section~\ref{#1}}

\newcommand{\fref}[1]{Fig.~\ref{#1}}
\newcommand{\sref}[1]{Sec.~\ref{#1}}
\newcommand{\eg}{\textit{e.g.}}
\newcommand{\ie}{\textit{i.e.}}

\newcommand{\hmm}[1]{{#1}}
\newcommand{\revision}[1]{{#1}}

\title{Deep Exemplar-based Colorization\thanks{Supplemental material: \url{http://www.dongdongchen.bid/supp/deep_exam_colorization/index.html}}}

\author{Mingming He${^\ast}^1$, Dongdong Chen${^\ast}^2$ \thanks{$^\ast$ indicates equal contribution. This work was done when Mingming He and Dongdong Chen were interns at Microsoft Research Asia.}, Jing Liao$^3$, 
	Pedro V. Sander$^1$, and Lu Yuan$^3$
	\\
	$^1$Hong Kong UST,  
	$^2$University of Science and Technology of China,
	$^3$Microsoft Research
}

\pdfauthor{He$^\ast$, Chen$^\ast$, Liao, Sander and Yuan}


\begin{document}

% A "teaser" figure, centered below the title and authors and above the body of the work.
\teaser{
  \centering
  \footnotesize
\includegraphics[width=1.0\linewidth]{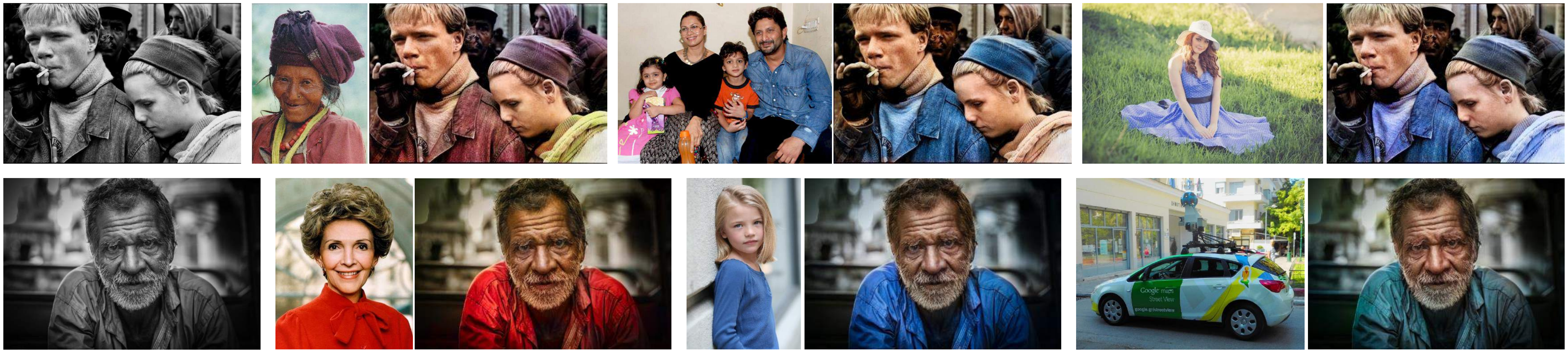}\\
\begin{tabular}{lllllll}
\qquad Target \qquad \qquad & Reference 1 \qquad & \qquad Result 1 \qquad \qquad &  Reference 2 & \qquad \quad Result 2 \qquad \qquad &  \qquad Reference 3 \qquad \qquad & \qquad Result 3 \qquad\\ 
\end{tabular}
\vspace*{-.1in}
\caption{Colorization results of black-and-white photographs. Our method provides the capability of generating multiple plausible colorizations by giving different references. \revision{Input images (from left to right, top to bottom): Leroy Skalstad/pixabay, Peter van der Sluijs/wikimedia, Bollywood Hungama/wikimedia, Lorri Lang/pixabay, Aamir Mohd Khan/pixabay, Official White House Photographer/wikimedia, Anonymous/wikimedia and K. Krallis/wikimedia.}}
\label{fig:teaser}
}

% Processes all of the front-end information and starts the body of the work.
\maketitle

% abstract
\begin{abstract}
	We propose the first deep learning approach for exemplar-based \hmm{local} colorization. Given a reference color image, our convolutional neural network directly maps a grayscale image to an output colorized image. Rather than using hand-crafted rules as in traditional exemplar-based methods, our end-to-end \hmm{colorization} network learns how to {\em select}, {\em propagate}, and {\em predict} colors from the large-scale data. The approach performs robustly and generalizes well even when using reference images that are unrelated to the input grayscale image. More importantly, as opposed to other learning-based colorization methods, our network allows the user to achieve customizable results by simply feeding different references. In order to further reduce manual effort in selecting the references, the system automatically recommends references with our proposed image retrieval algorithm, which considers both semantic and luminance information. The colorization can be performed fully automatically by simply picking the top reference suggestion. Our approach is validated through a user study and favorable quantitative comparisons to the-state-of-the-art methods. Furthermore, our approach can be naturally extended to video colorization. Our code and models will be freely available for public use.
\end{abstract}

\section{Introduction}
The aim of image colorization is to add colors to a gray image such that the colorized image is perceptually meaningful and visually appealing. The problem is ill-conditioned and inherently ambiguous since there are potentially many colors that can be assigned to the gray pixels of an input image (\eg, leaves may be colored in green, yellow, or brown). Hence, there is no unique correct solution and human intervention often plays an important role in the colorization process.

Manual information to guide the colorization is generally provided in one of two forms: user-guided scribbles or a sample reference image. In the first paradigm~\cite{levin2004colorization,yatziv2006fast,huang2005adaptive,luan2007natural,qu2006manga}, the manual effort involved in placing the scribbles and the palette of colors must be chosen carefully in order to achieve a convincing result. This often requires both experience and a good sense of aesthetics, thus making it challenging for an untrained user. In the second paradigm~\cite{welsh2002transferring,ironi2005colorization,tai2005local,charpiat2008automatic,liu2008intrinsic,chia2011semantic,gupta2012image,bugeau2014variational}, a color reference image similar to the grayscale image is given to facilitate the process. First, correspondence is established, and then colors are propagated from the most reliable correspondences. However, the quality of the result depends heavily on the choice of reference. Intensity disparities between the reference and the target caused by lighting, viewpoint, and content dissimilarity can mislead the colorization algorithm.

A more reliable solution is to leverage a huge reference image database to search for the most similar image patch/pixel for colorization. Recently, deep learning techniques have achieved impressive results in modeling large-scale data. Image colorization is formulated as a regression problem and deep neural networks are used to directly solve it~\cite{cheng2015deep,deshpande2015learning,larsson2016learning,iizuka2016let,zhang2016colorful,isola2016image,zhang2017real}. These methods can colorize a new photo fully automatically without requiring any scribbles or reference. Unfortunately, none of these methods allow multi-modal colorization \hmm{\cite{charpiat2008automatic}}. By learning from the data, their models mainly use the dominant colors they have learned, hindering any kind of user controllability. Another drawback is that it must be trained on a very large reference image database containing all potential objects.

More recent works attempt to achieve the best of both worlds: controllability from interaction and robustness from learning. Zhang et al.~\shortcite{zhang2017real} and Sangkloy et al.~\shortcite{sangkloy2016scribbler} add manual hints in the form of color points or strokes to the deep neural network in order to suggest possibly desired colors for the scribbles provided by users. This greatly facilitates traditional scribble-based interactions and achieves impressive results with more natural colors learned from the large-scale data. However, the scribbles are still essential for achieving high quality results, so a certain amount of trial-and-error is still involved.

In this paper, we suggest another type of hybrid solution. We propose the first deep learning approach for exemplar-based \hmm{local} colorization. Compared with existing colorization networks~\cite{cheng2015deep,iizuka2016let,zhang2016colorful}, our network allows control over the output colorization by simply choosing different references. As shown in \fref{fig:teaser}, the reference can be similar or dissimilar to the target, but we can always obtain plausible colors in the results, which are visually faithful to the references and perceptually meaningful.

To achieve this goal, we present the first convolutional neural network (CNN) to directly select, propagate and predict colors from an aligned reference for a gray-scale image. Our approach is qualitatively superior to existing exemplar-based methods. The success comes from two novel sub-networks in our exemplar-based colorization \hmm{framework}.

First, the \emph{Similarity sub-net} \hmm{is a pre-processing step which provides the input of the end-to-end colorization network. It} measures the semantic similarity between the reference and the target using a VGG-19 network pre-trained on the gray-scale image object recognition task. It provides a more robust and reliable similarity metric to varying semantic image appearances than previous metrics based on low-level features.

\begin{figure}
\centering
\footnotesize
	\includegraphics[width=1.0\linewidth]{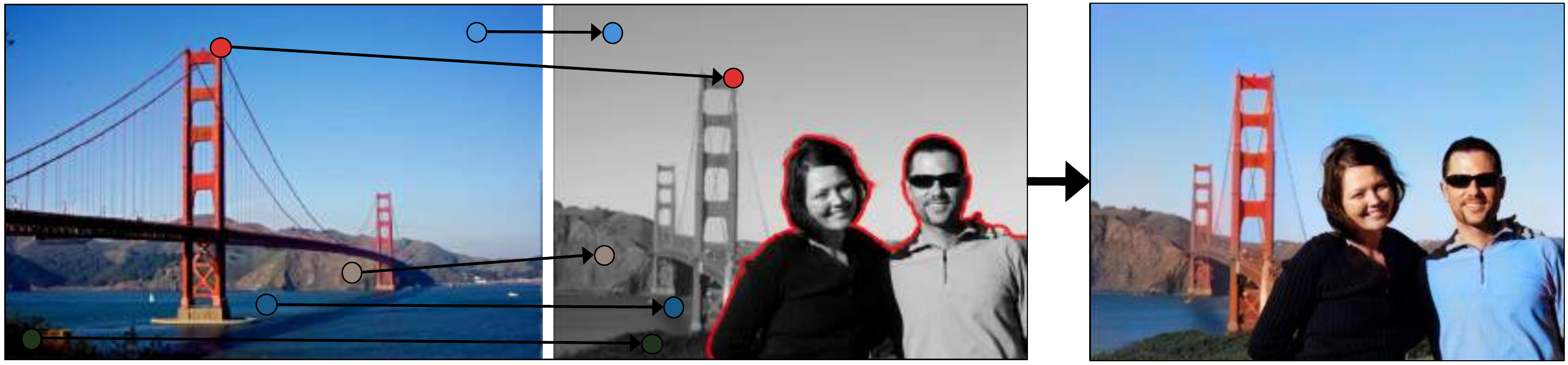}		
    \begin{tabular}{lll}
\quad    \qquad   Reference  \qquad \qquad   &    \qquad  Target  \qquad \qquad    &   \qquad   Colorized Result 
    \end{tabular}
    \vspace*{-.1in}
	\caption{Our goal is to selectively propagate the correct reference colors (indicated by the dots) for the relevant patches/pixels, and predict natural colors learned from the large-scale data when no appropriate matching region is available in the reference (indicated by the region outlined in red). \revision{Input images (from left to right): Julian Fong/flickr and Ernest McGray, Jr/flickr.}}
	\label{fig:intuition}
	\vspace{-1.5em}
\end{figure}

Then, the \emph{Colorization sub-net} provides a more general colorization solution for either similar or dissimilar patch/pixel pairs. It employs multi-task learning to train two different branches, which share the same network and weight but are associated with two different loss functions: 1) \emph{\hmm{Chrominance loss}}, which encourages the network to selectively propagate the correct reference colors for relevant patch/pixel, satisfying chrominance consistency; 2) \emph{Perceptual loss}, which enforces a close match between the result and the true color image of high-level feature representations. This ensures a proper colorization learned from the large-scale data even in cases where there is no appropriate matching region in the reference (see \fref{fig:intuition}). Therefore, our method can greatly loosen restrictive requirements on a good reference selection as required in other exemplar-based methods.

To guide the user towards efficient reference selection, the system recommends the most likely reference based on a proposed image retrieval algorithm. It leverages both high-level semantic information and low-level luminance statistics to search for the most similar images in the ImageNet dataset~\cite{russakovsky2015imagenet}. With the help of this recommendation, our approach can serve as a fully automatic colorization system. The experiments demonstrate that our automatic colorization outperforms existing automatic methods quantitatively and qualitatively, and even produces comparably high quality results to the-state-of-the-art interactive methods~\cite{zhang2017real,sangkloy2016scribbler}. Our approach can also be extended to video colorization.

Our contributions are as follows: (1) The first deep learning approach for exemplar-based colorization, which allows controllability and is robust to reference selection. (2) A novel end-to-end double-branch network architecture which jointly learns faithful \hmm{local} colorization to a meaningful reference and plausible color prediction when a reliable reference is unavailable. (3) A reference image retrieval algorithm for reference recommendation, with which we can also attain a fully automatic colorization. (4) A method capable of transferability to unnatural images, even though the network is trained purely on a natural image dataset. (5) An extension to video colorization.

\begin{figure*}[t]
	\centering
	\includegraphics[width=0.95\linewidth]{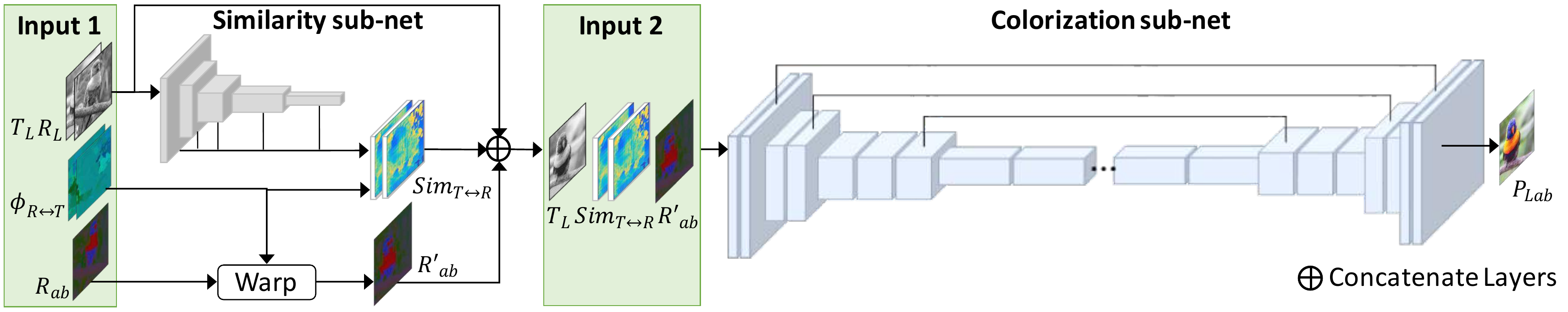}		
	\caption{System pipeline (inference stage). \hmm{The system consists of two sub-networks. The \emph{Similarity sub-net} works as a pre-processing step using Input 1 which includes two luminance channels $T_L$ and $R_L$ from the target and the reference respectively, bidirectional mapping functions $\phi_{T \leftrightarrow R}$ and \revision{two chrominance channels} $R_{ab}$ from the reference. It computes the bidirectional similarity maps $sim_{T \leftrightarrow R}$ and the aligned reference chrominance $R'_{ab}$, which, along with $T_L$, form Input 2 for the \emph{Colorization sub-net}. The \emph{Colorization sub-net} is an end-to-end CNN to predict the chrominance channels of the target, which are then combined with $T_L$ to generate the final colorized result $P_{Lab}$.}}
	\label{fig:sys}
	\vspace{-1.5em}
\end{figure*}

\section{Related work}
Next, we provide an overview of the major related works of each of the major algorithm categories.

\subsection{Scribble-based colorization}
These methods focus on propagating local user hints, for instance, color points or strokes, to the entire gray-scale image. The color propagation is based on some low-level similarity metrics. The pioneering work of Levin et al.~\shortcite{levin2004colorization} assumed that adjacent pixels with similar luminance should have similar color, and then solved a Markov Random Field for propagating sparse scribble colors. Further advances extended similarity to textures~\cite{qu2006manga,luan2007natural}, intrinsic distance~\cite{yatziv2006fast}, and exploited edges to reduce color bleeding~\cite{huang2005adaptive}. The common drawback of such methods is intensive manual work and professional skills for providing good scribbles.	

\subsection{Example-based colorization}
These methods provide a more intuitive way to reduce extensive user effort by feeding a very similar reference to the input grayscale image. The earliest work~\cite{welsh2002transferring} transferred colors by matching global color statistics, similar to Reinhard et al.~\shortcite{reinhard2001color}. The approach yielded unsatisfactory results in many cases since it ignored spatial pixel information. For more accurate local transfer, different correspondence techniques are considered, including segmented region level~\cite{ironi2005colorization,tai2005local,charpiat2008automatic}, super-pixel level~\cite{gupta2012image,chia2011semantic}, and pixel level~\cite{liu2008intrinsic,bugeau2014variational}. However, finding low-level feature correspondences (\eg, SIFT, Gabor wavelet) with hand-crafted similarity metrics is susceptible to error in situations with significant intensity and content variation. Recently two works utilize deep features extracted from a pre-trained VGG-19 network for reliable matching between images that are semantically-related but visually different, and then leverage it to style transfer~\cite{liao2017visual} and color transfer~\cite{he2017neural}. However, all of these exemplar-based methods have to rely on finding a good reference, which is still an obstacle for users, even when some semi-automatic retrieval methods~\cite{liu2008intrinsic,chia2011semantic} are used. By contrast, our approach is robust to any given reference thanks to the capability of \hmm{our} deep network to learn natural color distributions from large-scale image data.

\subsection{Learning-based colorization}
Several techniques rely entirely on learning to produce the colorization result. Deshpande et al.~\shortcite{deshpande2015learning} defined colorization as a linear system and learned its parameters. Cheng et al.~\shortcite{cheng2015deep} concatenated several pre-defined features and fed them \hmm{into} a three-layer fully connected neural network. Recently, some end-to-end learning approaches~\cite{larsson2016learning,iizuka2016let,zhang2016colorful,isola2016image} leveraged CNN to automatically extract features and predict the color result. The key difference in those networks is the loss function (\eg\hmm{,} image reconstruction $L_2$ loss~\cite{iizuka2016let}, classification loss~\cite{larsson2016learning,zhang2016colorful}, and $L_1 + GAN$ loss for considering the multi-modal colorization~\cite{isola2016image}). All of these networks are learned from large-scale data and do not require any user intervention. However, they only produce a single plausible result for each input, even though colorization is intrinsically an ill-posed problem with multi-modal uncertainty~\cite{charpiat2008automatic}.

\subsection{Hybrid colorization}
To achieve desirable color results, Zhang et al.~\shortcite{zhang2017real} and Sangkloy et al.~\shortcite{sangkloy2016scribbler} proposed a hybrid framework that inherits the controllability from scribble-based methods and the robustness from learning-based methods. Zhang et al.~\shortcite{zhang2017real} uses provided color points while Sangkloy et al.~\shortcite{sangkloy2016scribbler} adopts strokes. Instead, we incorporate the reference rather than user-guided points or strokes into the colorization network, since we believe that giving a similar color example is \hmm{a} more intuitive way for untrained users. Furthermore, the reference selection can be achieved automatically using our image retrieval system.

\section{Exemplar-based Colorization Network}

Our goal is to colorize a target grayscale image based on a color reference image. More specifically, we aim to apply a reference color to the target where there is semantically-related content, and fall back to a plausible colorization for the objects or regions with no related content in the reference. To achieve this goal, we address two major challenges.

First, it is difficult to measure the semantic relationship between the reference and the target, especially given that the reference is in color while the target is a grayscale image. To solve this problem, we use a gray-VGG-19, trained on image classification tasks only using the luminance channel to extract their own features, and compute their feature's differences.

Second, it is still challenging to select reference colors and propagate them properly by defining hand-crafted rules based on the similarity metrics. Instead, we propose an end-to-end network to learn selection and propagation simultaneously. Oftentimes both steps are not enough to recover all colors, especially when the reference
is not very related to the target. To address this issue, our network would instead predict the dominant colors for misaligned objects from the large-scale data.

Fig. \ref{fig:sys} illustrates the system pipeline. Our system uses the CIE Lab color space, which is perceptually linear. Thus, each image can be separated into a luminance channel $L$ and two chrominance channels $a$ and $b$. The input of our system includes a grayscale target image $T_L \in \mathcal{R}^{H\times W \times 1}$, a color reference image $R_{Lab} \in \mathcal{R}^{H\times W \times 3}$, and the bidirectional mapping functions between them. \hmm{The bidirectional mapping function is a spatial warping function defined with bidirectional correspondences. It returns the transformed pixel location given a source location "p". The two functions are respectively denoted as $\phi_{T \rightarrow R}$ (mapping pixels from $T$ to $R$) and $\phi_{R \rightarrow T}$ (mapping pixels from $R$ to $T$), where $H$ and $W$ are the height and width of the input images.} For simplicity, we assume the two input images are of the same dimensions, although this is not necessary in practice. Our network consists of two sub-networks. The \emph{Similarity sub-net} computes the semantic similarities between the reference and the target, and outputs bidirectional similarity maps $sim_{T \leftrightarrow R}$. The \emph{Colorization sub-net} takes $sim_{T \leftrightarrow R}$, $T_L$ and $R'_{ab}$ as the input, and outputs the predicted $ab$ channels of the target $P_{ab} \in \mathcal{R}^{H\times W \times 2}$, which \hmm{are} then combined with $T_L$ to get the colorized result $P_{Lab}$ ($P_L=T_L$). Details of \hmm{the} two sub-networks are introduced in the following sections.

\subsection{Similarity Sub-Network}
\label{sec:simnet}

Before calculating pixel-level similarity, the two input images $T_L$ and $R_{Lab}$ have to be aligned. The bidirectional mapping functions $\phi_{R \rightarrow T}$ and $\phi_{T \rightarrow R}$ can be calculated with a dense correspondence algorithm, such as SIFTFlow~\cite{liu2011sift}, Daisy Flow~\cite{tola2010daisy} or DeepFlow~\cite{weinzaepfel2013deepflow}. In our work, we adopt the latest advanced technique called Deep Image Analogy~\cite{liao2017visual} for dense matching, since it is capable of matching images that are visually different but semantically related.

Our work is inspired by recent observations that CNNs trained on image recognition tasks are capable of encoding a full spectrum of features, from low-level textures to high-level semantics. It provides \hmm{a} robust and reliable similarity metric to variant image appearances (caused by variant lightings, times, viewpoints, and even slightly \hmm{different} categories), which may be challenging for low-level feature metrics (\eg, intensity, SIFT, Gabor wavelet) used in many works~\cite{welsh2002transferring,liu2008intrinsic,charpiat2008automatic,tai2005local}.

\begin{table}[t]
\caption{Classification accuracies of original and our fine-tuned VGG-19 calculated on ImageNet validation dataset.}
 \vspace*{.025in}
 \begin{tabular}{lcc}
  \toprule
      & Top-5 Class & Top-1 Class \\
      & Acc($\%$) & Acc($\%$)\\
  \midrule
 Ori VGG-19 tested on color image & 91.24 & 73.10 \\
 Ori VGG-19 tested on gray image & 83.63 & 61.14\\
 Our VGG-19 tested on gray image & \textbf{89.39} & \textbf{70.05}\\
  \bottomrule
 \end{tabular} 
 %\caption{Classification accuracies of original and our fine-tuned VGG-19 calculated on ImageNet validation dataset.}
 \label{tab:acc}
 \end{table}

We take the intermediate output of VGG-19 as our feature representation. Certainly, other recognition networks, such \hmm{as} GoogleNet~\cite{Szegedy2015googlenet} or ResNet~\cite{He2015resnet} can also be used. The original VGG-19 is trained on color images and has a degraded accuracy of recognizing grayscale images, as shown in \Tref{tab:acc}. To reduce the performance gap between color images and their gray versions, we train a gray-VGG-19 only using \hmm{the} luminance channel of an image. It increases the top-5 accuracy from $83.63\%$ to $89.39\%$, and approaches that of \hmm{the} original VGG-19 ($91.24\%$) evaluated on color images.

We then feed the two luminance channels $T_L$ and $R_L$ into our gray-VGG-19 respectively, and obtain their five-level feature map pyramids ($i = 1...5$). The feature map of each level $i$ is extracted from the $relu\{i\}\_1$ layer. Note that the features have progressively coarser spatial resolution with increasing levels. We upsample all feature maps to the same spatial resolution of the input images and denote the upsampled feature maps of $T_L$ and $R_L$ as $\{F_{T_L}^i\}_{i=1,...,5}$ and $\{F_{R_L}^i\}_{i=1,...,5}$ respectively. Bidirectional similarity maps $sim_{T \rightarrow R}^i$ and $sim_{R \rightarrow T}^i$ are computed between $F_{T}^i$ and $F_{R}^i$ at each pixel $p$:
\begin{equation}
\begin{split}
sim_{T \rightarrow R}^i(p) &= d(F_{T}^i(p), F_{R}^i(\phi_{T \rightarrow R}(p))\ ,\\
sim_{R \rightarrow T}^i(p) &= d(F_{T}^i(\phi_{R \rightarrow T}(\phi_{T \rightarrow R}(p))), F_{R}^i(\phi_{T \rightarrow R}(p)))\ .
\end{split}
\end{equation}
\hmm{As mentioned in Liao et al.~\shortcite{liao2017visual}, cosine distance performs better in measuring feature similarity since it is more robust to appearance variances between image pairs.} Thus, our similarity metric $d(x,y)$ between two deep features is defined as their cosine similarity:
\begin{equation}
d(x,y)=\frac{x^{\hmm{T}}y}{|x||y|}\ .
\label{eq:cos}
\end{equation}

The forward similarity map $sim_{T \rightarrow R}$ reflects the matching confidence from $T_L$ to $R_L$ while the backward similarity map $sim_{R \rightarrow T}$ measures the matching accuracy in the reverse direction. We use $sim_{T \leftrightarrow R}$ to denote both.

\begin{figure}[t]
	\centering
	\includegraphics[width=0.95\linewidth]{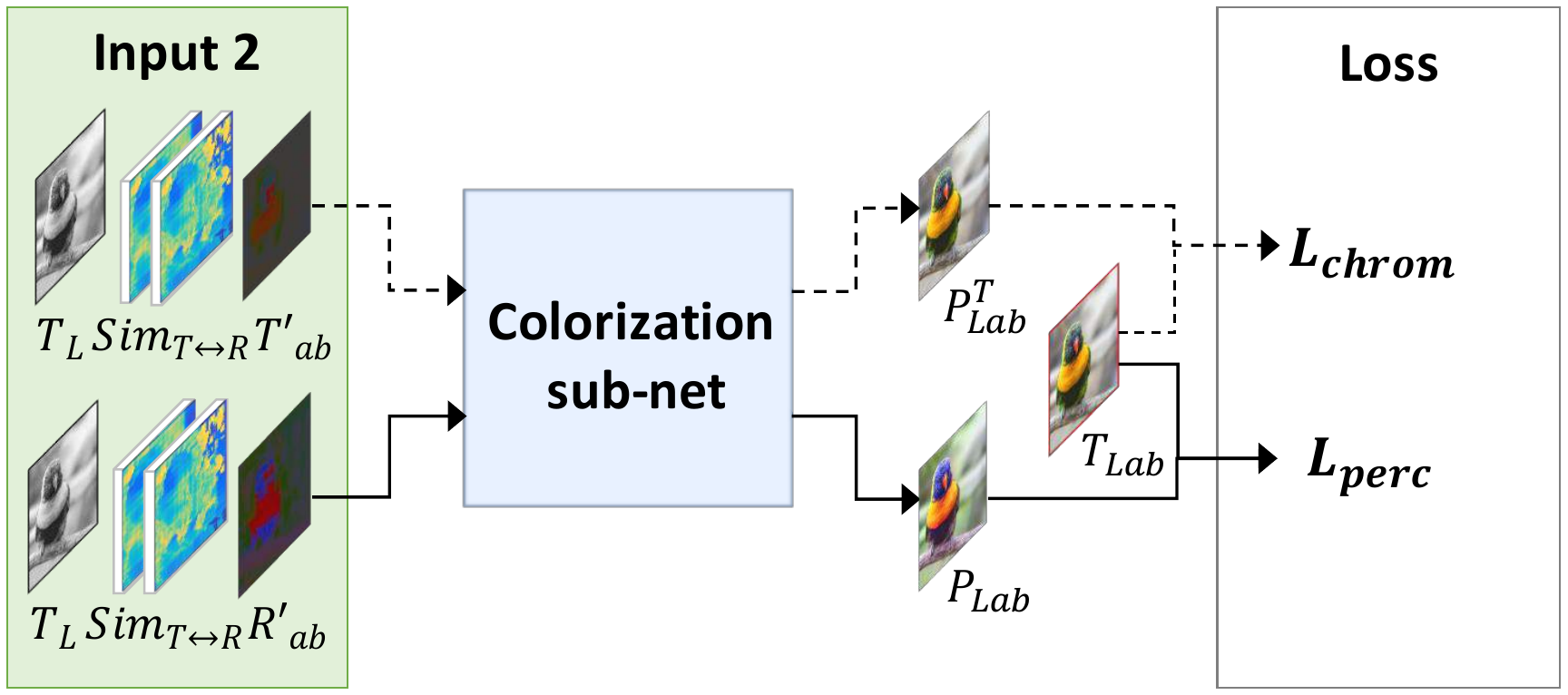}	
	\caption{Two branches training of \emph{Colorization sub-net}. \hmm{Both branches take nearly the same Input 2 except for the concatenated chrominance channel. The aligned ground truth chrominance $T'_{ab}$ is used for the \emph{\hmm{Chrominance branch}} to compute the chrominance loss $L_{chrom}$, while the aligned reference chrominance $R'_{ab}$ is used in the \emph{Perceptual branch} to compute the perceptual loss $L_{perc}$.}}
	\label{fig:loss}
	\vspace{-1.5em}
\end{figure}

\subsection{Colorization Sub-Network}
\label{sec:col}

We design an end-to-end CNN $\mathcal{C}$ to learn selection, propagation and prediction of colors simultaneously. As shown on the right of \fref{fig:sys}, $\mathcal{C}$ takes a \hmm{thirteen}-channel map as the input, which concatenates the gray target $T_L$, aligned reference with chrominance channels only $R'_{ab}(p)=R_{ab}(\phi_{T\rightarrow R}(p))$, and bidirectional similarity maps $sim_{T \leftrightarrow R}$ ). It also predicts $ab$ channels of the target image $P_{ab}$. Next, we describe the loss function, network architecture and training strategy of the network.

\subsubsection{Loss}
Usually, the objective of colorization is to encourage the output $P_{ab}$ to be as close as possible to the ground truth $T_{ab}$, the original $ab$ channels of a color image $T_{Lab}$ in the training dataset. However this is not true in exemplar-based colorization, because the colorization $P_{ab}$ should allow customization with $R'_{ab}$ (\eg, a flower can be colorized in either red, yellow, purple depending on the reference). Thus, it is not accurate to directly penalize a measure of the difference between $P_{ab}$ and $T_{ab}$, as in other colorization networks (\eg, using $L_2$ loss~\cite{cheng2015deep,iizuka2016let}, $L_1$ loss~\cite{isola2016image,zhang2017real}, or classification loss~\cite{larsson2016learning,zhang2016colorful}).

Instead, our objective function is designed to consider two desiderata. First, we prefer reliable reference colors to be applied in the output, thus making it faithful to the reference. Second, we encourage the colorization to be natural, even when no reliable reference color is available.

To achieve both goals, we propose a multi-task network, which involves two branches, \emph{\hmm{Chrominance branch}} and \emph{Perceptual branch}. Both branches share the same network $\mathcal{C}$ and weight $\theta$ but are associated with their own input and loss functions, as shown in \fref{fig:loss}. A parameter $\alpha$ is used to dictate the relative weight between the two branches.

In the \emph{\hmm{Chrominance branch}}, the network learns to selectively propagate the correct reference colors, which depends on how well the target $T_L$ and the reference $R_L$ are matched. \hmm{However, training such a network is not easy: 1) on the one hand, the network cannot be trained directly with $R'_{ab}$, the reference chrominance warped on the target, because the corresponding ground truth colorization is unknown; 2) while on the other hand, the network cannot be trained using the ground truth target chrominance $T_{ab}$ as a reference, because that would essentially be providing the network with the answer it is supposed to predict. Thus, we leverage the bidirectional mapping functions to reconstruct a "fake" reference $T'_{ab}$ from the ground truth chrominance, \ie, $T'_{ab}(p) = T_{ab}(\phi_{R \rightarrow T}(\phi_{T \rightarrow R}(p))$. $T'_{ab}$ replaces $R'_{ab}$ in the training stage with the underlying hypothesis that correct color samples in $T'_{ab}$ are very likely to lie in the same positions as correct color samples in $R'_{ab}$, since both are warped with 
$\phi_{T \rightarrow R}$. }

To train the chrominance branch, both $T_{L}$ and $T'_{ab}$ are fed to the network, yielding the result $P^T_{ab}$:
\begin{align}
P^T_{ab}=\mathcal{C}(T_{L},sim_{T \leftrightarrow R},T'_{ab};\theta)\ .
\end{align}
Here, $P^T_{ab}$ is colorized with the guidance of $T'_{ab}$, and should recover the ground truth $T_{ab}$ if the network selects the correct color samples and propagates them properly. The smooth $L_1$ distance is evaluated at each pixel $p$ and integrated over the entire image to evaluate the \emph{\hmm{Chrominance loss}}:
\begin{align}
\mathcal{L}_{chrom}(P^T_{ab})=\sum\nolimits_{p}{smooth\_L_{1}(P^T_{ab}(p),T_{ab}(p))}\,
\label{eq:cycle_loss}
\end{align}
where $smooth\_L_{1}(x,y)=\frac{1}{2}(x-y)^2$, if $|x-y|<1$, $smooth\_L_{1}(x,y)= |x-y|-\frac{1}{2}$, otherwise. We take the smooth $L_1$ loss as the distance metric to avoid the averaging solution in the ambiguous colorization problem~\cite{zhang2017real}.

Using the {\em \emph\hmm{Chrominance branch}} only works for reliable color samples in $R'_{ab}$ but may fail when the reference is dissimilar to parts of the image. To allow the network to predict perceptually plausible colors even without a proper reference, we add a {\em Perceptual branch}. In this branch, we take the reference $R'_{ab}$ and the target $T_L$ as the network input during training. Then, we generate the predicted chrominance $P_{ab}$:
\begin{align}
P_{ab}=\mathcal{C}(T_L,sim_{T \leftrightarrow R},R'_{ab};\theta)\ .
\end{align}

\begin{figure}[t]
	\footnotesize
	\setlength{\tabcolsep}{0.007\linewidth}
	\begin{tabular}{ccc}
      	\includegraphics[height=0.25\linewidth]{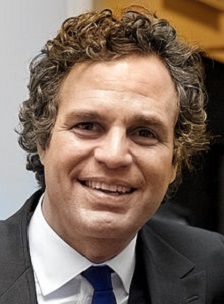}&
		\includegraphics[height=0.25\linewidth]{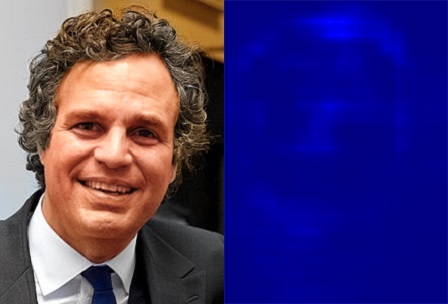}&
		\includegraphics[height=0.25\linewidth]{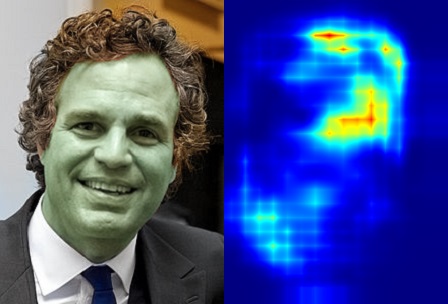} \\
		Ground truth & Colorized result 1 $\&$ & Colorized result 2 $\&$ \\
		 & Error map & Error map
	\end{tabular}
    \vspace*{-.1in}
	\caption{Visualization of \emph{Perceptual loss}. Both colorized results have the same $L_2$ chrominance ({\em{ab}} channels) distance to the ground truth, but the unnatural green face (right) has a much larger \emph{Perceptual loss} than a more plausible skin color (left). \revision{Input image: Zhang et al.~\protect\shortcite{zhang2017real}.}}
	\label{fig:err}
	\vspace{-1.5em}
\end{figure}

In this branch, we minimize \emph{Perceptual loss}~\cite{johnson2016perceptual} instead. Formally:
\begin{equation}
\mathcal{L}_{perc}(P_{ab})=\sum\nolimits_{p}{||{F}_P(p)-{F}_T(p)||^2}\,
\label{eq:perc_loss}
\end{equation}
where ${F}_P$ represents the feature maps extracted from the original VGG19 $relu5\_1$ layer for $P_{Lab}$, and ${F}_T$ is the same for $T_{Lab}$. \emph{Perceptual loss} measures the semantic differences caused by unnatural colorization and is robust to appearance differences caused by two plausible colors, as shown in \fref{fig:err}. \hmm{We also did some exploration using cosine distance but found L2 distance generated superior results.} A similar loss is widely used in other tasks, like style transfer~\cite{chen2017coherent,chen2018stereoscopic}, photo-realistic image synthesis~\cite{chen2017photographic}, and super resolution~\cite{sajjadi2017enhancenet}.

Our network $\mathcal{C}$, parameterized by $\theta$, learns to minimize both loss functions (\Eref{eq:cycle_loss} and \eqref{eq:perc_loss}) across a large dataset:
\begin{equation}
\theta^*=\mathop{\arg\min}_{\theta}(\mathcal{L}_{chrom}(P^T_{ab})+\alpha \mathcal{L}_{perc}(P_{ab})),
\label{eq:loss}
\end{equation}
where $\alpha$ is empirically set to $0.005$ to balance both branches.

\begin{figure*}[t]
	\centering
	\includegraphics[width=0.9\linewidth]{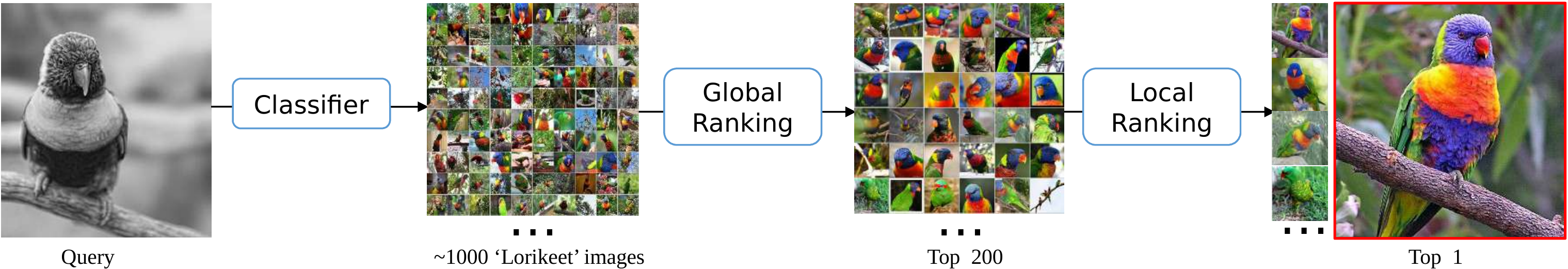}	
    \vspace*{-.1in}
	\caption{Color reference recommendation pipeline. \revision{Input images: ImageNet dataset.}}
	\label{fig:retrival}
	\vspace{-1.5em}
\end{figure*}

\subsubsection{Architecture}\label{sec:arch} 
The sub-network adopts a U-net encoder-decoder structure with some skip connections between the lower layers and symmetric higher layers. \hmm{We empirically chose the U-net architecture because of its effectiveness, as evidenced in many image generation tasks~\cite{badrinarayanan2015segnet,yu2015multi,zhang2017real}.} Specifically, our network consists of 10 convolutional blocks. Each convolutional block contains $2\sim3$ $conv$-$relu$ pairs, followed by a batch normalization layer~\cite{ioffe2015batch} with the exception of the last block. The feature maps in \hmm{the} first $4$ convolutional blocks are progressively halved spatially while doubling the feature channel number. To aggregate multi-scale contextual information without losing resolution (as in Yu et al.~\shortcite{yu2015multi}, Zhang et al.~\shortcite{zhang2017real} and Fan et al.~\shortcite{fan2018decouple}), dilated convolution layers with a factor of $2$ are used in the $5th$ and $6th$ convolutional blocks. In the last $4$ convolutional blocks, feature maps are progressively doubled spatially while halving the feature channel number. All down-sampling layers use convolution with stride $2$, while all up-sampling layers use deconvolution with stride $2$. Symmetric skip connections are added between the outputs of $1st$ and $10th$, $2nd$ and $9th$, and $3rd$ and $8th$ blocks, respectively. Finally, a convolution layer with a kernel size $1 \times 1$ is added after the $10th$ block to predict the output $P_{ab}$. The final layer is \hmm{a} $tanh$ layer (also used in Radford et al.~\shortcite{radford2015unsupervised} and Chen et al.~{chen2017stylebank}), which makes $P_{ab}$ within a meaningful bound.

\subsubsection{Dataset} \label{sec:dataset}
We generate a training dataset based on ImageNet dataset~\cite{russakovsky2015imagenet} by sampling approximately 700,000 image pairs from 7 popular categories: animals ($15\%$), plants ($15\%$), people ($20\%$), scenery ($25\%$), food ($5\%$), transportation ($15\%$) and artifacts ($5\%$), involving 700 classes out of the total 1,000 classes \hmm{due to the cost of generating training data}. To let the network be robust to any reference, we sample image pairs with different extents of similarity. Specifically, $45\%$ of image pairs belong to Top-5 similarity (selected by our recommendation algorithm described in \Sref{sec:rec}) in the same class. Another $45\%$ are randomly sampled within the same class. The remaining $10\%$ have less similarity as they are randomly sampled from different classes but within the same category. In the training stage, we randomly switch the role of the two images for each pair to augment data. In other words, the target and the reference can be switched as two variant pairs during training. All images are scaled with the shortest edge of $256$ pixels.

\subsubsection{Training}
Our network is trained using the Adam optimizer~\cite{kingma2014adam} with a batch size of 256. \hmm{For every iteration, within the batch, the first $50\%$ of data (128) go through the \emph{\hmm{Chrominance branch}} using use $T'_{ab}$ as a reference and the remaining $50\%$ (128) go through the \emph{Perceptual branch} using $R'_{ab}$. The two branches respectively use corresponding losses.} When updating the \emph{\hmm{Chrominance branch}}, only \emph{\hmm{Chrominance loss}} is used for gradient back propagation. When updating the \emph{Perceptual branch}, only \emph{Perceptual loss} is used for gradient back propagation. The initial learning rate is set to $0.0001$ and decays by 0.1 every 3 epochs. By default, we train the whole network with 10 epochs. The whole training procedure takes around 2 days on 8 x Titan XP GPUs.	

\section{Color Reference Recommendation}
\label{sec:rec}
As discussed earlier, our network is robust to reference selection, and provides user control for the colorization. To aid users in finding good references, we propose a novel image retrieval algorithm that automatically recommends good references to the user. Alternatively, the approach yields a fully automatic system by directly using the Top-1 candidate.

The ideal reference is expected to match the target image in both semantic content and photometric luminance. The purpose of incorporating the luminance term is to avoid any unnatural composition of luminance and chrominance. In other words, combining the reference chrominance with the target luminance may produce visually unfaithful colors to the reference. Therefore, we desire the reference's luminance to be as close as possible to the target's.

To measure semantic similarity, we adopt the intermediate features of a pre-trained image classification network as descriptors, which have been widely used in recent image retrieval works \cite{krizhevsky2012imagenet,babenko2014neural,gong2014multi,babenko2015aggregating,Razavian2016baseline,tolias2015particular}. 

We propose an effective and efficient image retrieval algorithm. The system overview is shown in \fref{fig:retrival}. We feed the luminance channel of each image from our training dataset (see \Sref{sec:dataset}) to our pre-trained gray-VGG-19 (in \Sref{sec:simnet}), and get its feature $F^{5}$ from the last convolutional layer $relu5\_4$ and $F^{6}$ from the first fully-connected layer $fc6$. These features are pre-computed and stored in the database for the latter query. We also feed the query image (\ie, the target gray image) to the gray-VGG-19 network, and get its corresponding features $F_{T}^{5}, F_{T}^{6}$. We then proceed with two ranking steps described next.

\subsubsection{Global Ranking}
Through gray-VGG-19, we can also get the recognized Top-1 class ID for the query image $T_L$. According to the class ID, we narrow down the search domain to all the images ($\sim 1,000$ images) within the same class. Here, we want to further filter out dissimilar candidates by comparing $fc$ features between the query and all candidates. Even within the same class, the candidate could have a context that is irrelevant to the query. For example, the query could be "a cat running on grass", but the candidate could be "a cat sitting inside the house". We would like the semantic content in the two images to be as similar as possible \hmm{however}. To achieve this, for each candidate image $R_i (i=1,2,3,...)$ in this class, we directly compute the cosine similarity (in \Eref{eq:cos}) between $F_T^6$ and $F_{R_i}^6$ as the global score and rank all candidates by their scores.

\begin{figure*}
\footnotesize
\setlength{\tabcolsep}{0.003\linewidth}
\begin{tabular}{ccccccc}
\includegraphics[width=0.136\textwidth]{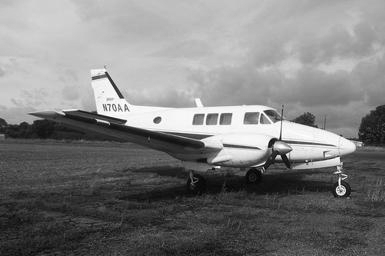}&
\includegraphics[width=0.136\textwidth]{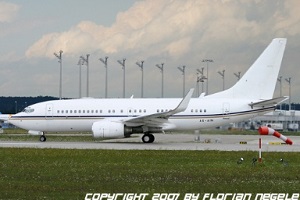}&
    \includegraphics[width=0.136\textwidth]{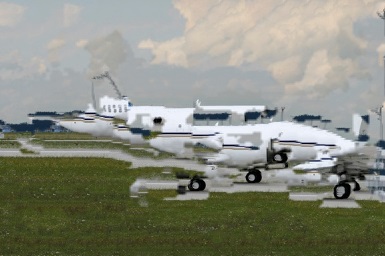}&
    \includegraphics[width=0.136\textwidth]{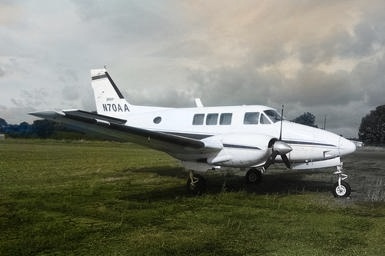}&
   \includegraphics[width=0.136\textwidth]{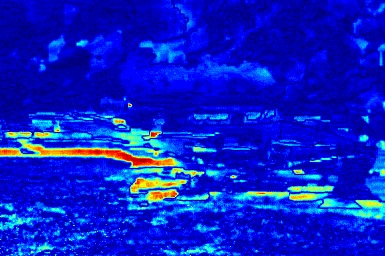}&
    \includegraphics[width=0.136\textwidth]{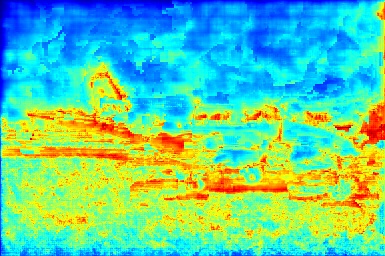}&
    \includegraphics[width=0.136\textwidth]{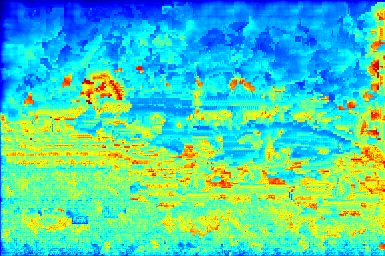}
\\
Target & Reference & Aligned reference & Predicted result & Chrominance 
difference & Matching error & Matching error \\
$T_L$ & $R$ & $R'$ & $ T_L \bigoplus R'_{ab}$ & $ |P_{ab}-R'_{ab}|$ & $1-sim_{T\rightarrow R} $&$1-sim_{R\rightarrow T} $
 \end{tabular}
  \vspace*{-.1in}
 \caption{Visualization of color selection in the \emph{\hmm{Chrominance branch}}. The points with smaller difference between the predicted colorization $P_{ab}$ and aligned reference color $R'_{ab}$ are most likely to be selected by the network and maintained in the final results. Note how inconsistencies between the similarity maps and the true color difference make it difficult to determine good points by the hand-crafted rules. \revision{Input images: ImageNet dataset.}}
  \label{fig:abl1-2}
  \vspace{-0.5em}
\end{figure*}

\begin{figure*}
\footnotesize
\setlength{\tabcolsep}{0.003\linewidth}
\scalebox{0.99}{
\begin{tabular}{cccccccc}
\includegraphics[width=0.12\textwidth]{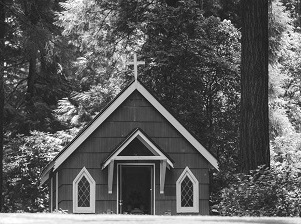}&
 \includegraphics[width=0.12\textwidth]{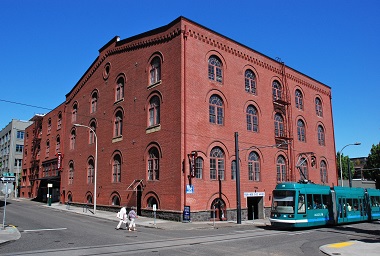}&
  \includegraphics[width=0.12\textwidth]{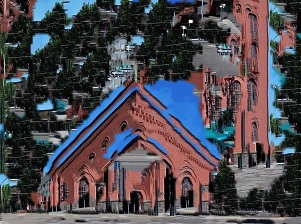}&
  \includegraphics[width=0.12\textwidth]{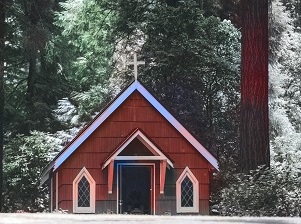}&
   \includegraphics[width=0.12\textwidth]{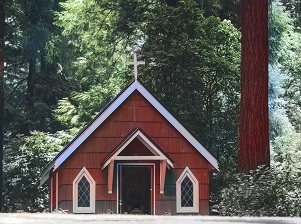}&
 \includegraphics[width=0.12\textwidth]{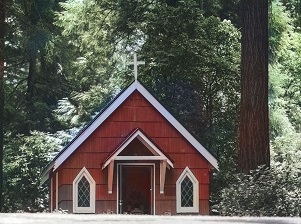}&
 \includegraphics[width=0.12\textwidth]{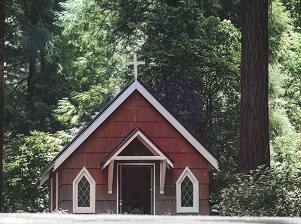}&
  \includegraphics[width=0.12\textwidth]{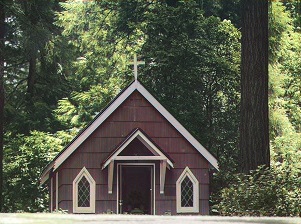}\\
 \includegraphics[width=0.12\textwidth]
 {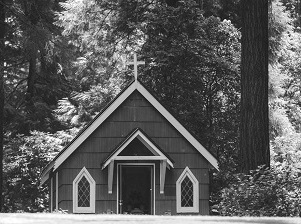}&
 \includegraphics[width=0.075\textwidth]{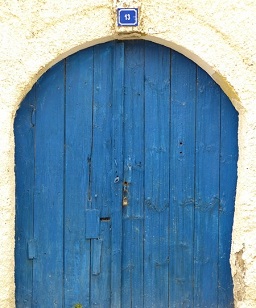}&
  \includegraphics[width=0.12\textwidth]{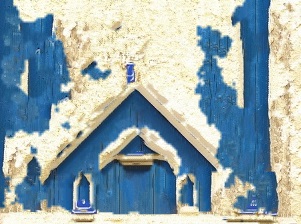}&
  \includegraphics[width=0.12\textwidth]{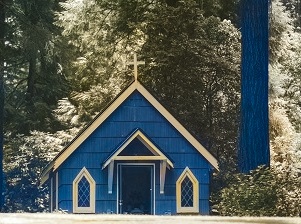}&
   \includegraphics[width=0.12\textwidth]{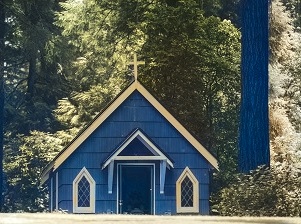}&
 \includegraphics[width=0.12\textwidth]{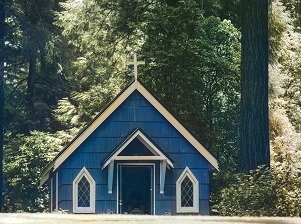}&
 \includegraphics[width=0.12\textwidth]{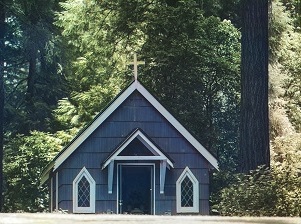}&
  \includegraphics[width=0.12\textwidth]{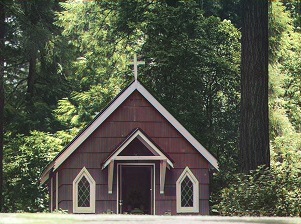}\\
   Target & Reference & Aligned reference & Chrominance  & Two branches & Two branches & Two branches & Perceptual  \\
   & & &branch only & $(\alpha=0.003)$ & $(\alpha=0.005)$ & $(\alpha=0.01)$ & branch only
 \end{tabular}}
 \vspace*{-.1in}
 \caption{Comparison of results from the training with different branch configurations. \revision{Input images (from left to right, top to bottom): Tabitha Mort/pexels, Steve Morgan/wikimedia and Anonymous/pxhere.}}
  \label{fig:abl1-1}
    \vspace{-0.8em}
\end{figure*}

\subsubsection{Local Ranking}
The global ranking provides us the top-$N$ (we set $N=200$) candidates $R_i$. As we know, $fc$ features fail to provide more accurate information about the object since it ignores the spatial information. For this purpose, we further prune these candidates by conducting a local ranking on the remaining $N$ images. The local similarity score consists of both semantic and luminance terms.

For each image pair $\{T_L, R_i\}$, at each point $p$ in $F_T^{5}$, we find its nearest neighbor $q$ in $F_{R_i}^{5}$ by minimizing the cosine distance between $F_T^{5}(p)$ and $F_{R_i}^{5}(q)$, namely $q = NN(p)$. Then, the semantic term is defined as the cosine similarity $d(\cdot)$ (see \Eref{eq:cos}) between two feature vectors $F_T^{5}(p)$ and $F_{R_i}^{5}(q)$. 

The luminance term measures the similarity of luminance statistics between two local windows corresponding to $p$ and $q$ respectively. We evenly split image $T_L$ into a 2D grid with each grid having $16 \times 16$ resolution. Each grid in the image $T_L$ indeed corresponds to a point in its feature map $F_T^{5}$, since it undergoes 4 down-sampling layers. $C_T(p)$ is denoted as the grid cell in the image corresponding to the point $p$ in $F_T^{5}$. Likewise, $C_{R_i}(q)$ from $R_L$ corresponds the point $q$ in $F_{R_i}^{5}$. The function $d_H(\cdot)$ measures the correlation coefficient between luminance histograms of $C_T(p)$ and $C_{R_i}(q)$.

The local similarity score is summarized as:
\begin{equation}
score(T, R_i) = \sum_{p} (d(F_T^{5}(p), F_{R_i}^{5}(q)) + \beta d_H(C_T(p), C_{R_i}(q))),
\end{equation}
where $\beta$ determines the relative importance between the two terms (empirically set to $0.25$). This similarity score is computed for each pair $\{T_L, R_i\} (i=1,2,3,...)$. According to all local scores, we re-rank all retained candidates and retrieve the top selections.

We \hmm{compress} neural features with the common PCA-based compression~\cite{babenko2014neural} \hmm{to accelerate the search.} The channels of feature $fc6$ are compressed from $4,096$ to $128$ and the channels of features $relu5\_4$ are reduced from $512$ to $64$ with practically negligible loss. After these dimensionality reductions, our reference retrieval can run in real-time.

\section{Discussion}

In this section, we analyze and demonstrate the capabilities of our colorization network through ablation studies.

\subsection{What does the \emph{Colorization sub-net} learn?}
\label{sec:whatlearn}

The \emph{Colorization sub-net} $\mathcal{C}$ learns how to select, propagate, and predict colors based on the target and the reference. As discussed earlier, it is an end-to-end network that involves two branches, each playing a distinct role. At first, we want to understand the behavior of the network using just the \emph{\hmm{Chrominance branch}} during learning. For this purpose, we only train the \emph{\hmm{Chrominance branch}} of $\mathcal{C}$ by minimizing the \emph{\hmm{Chrominance loss}} (in \Eref{eq:cycle_loss}), and evaluate it on one example to intuitively understand its operation (\fref{fig:abl1-2}). \hmm{By comparing the chrominance of the predicted result ($4th$ column) with the chrominance of the aligned reference ($3rd$ column), we notice that they have consistent colors in most regions (\eg, "blue" sky, "white" plane and "green" lawn).} That indicates that our \emph{\hmm{Chrominance branch}} picks color samples from the reference and propagates them to the entire image to achieve a smooth colorization.

To learn which color samples are selected by the network, we compute the chrominance difference between the predicted result and the aligned reference in the $5th$ column ("blue" denotes nearly no difference while "red" denotes a noticeable difference). Colors of the points with smaller errors are more likely to be selected by the network and then retained in the final result. 

"\emph{How does the network infer good samples?}" or "\emph{Can it be directly inferred from the matching between images?}" To answer these questions, we compare the difference map ($6th$ column) with the averaged five-levels matching errors $1 - sim_{T \rightarrow R}$ ($7th$ column) and $1 - sim_{R \rightarrow T}$ ($8th$ columns). On the one hand, we can see that the matching errors are essentially consistent with the difference. This demonstrates that our network can learn a good sampling based on the matching quality, which serves as a key "hint" to determine appropriate locations. On the other hand, we find that the network does not always select points with smaller matching errors, as evidenced by a significant number of inconsistent samples. \hmm{Without similarity maps, the Colorization sub-net can hardly infer the matching accuracy between the aligned reference and the input. It will also increase ambiguity of the color prediction.
} Thus, adaptive selection according to similarities may be infeasible through an intuitive heuristic. However, by using the large-scale data, our network can more robustly learn this mechanism directly. 

To understand the role of the \emph{Perceptual branch}, we train it by solely minimizing the \emph{Perceptual loss} (in \Eref{eq:perc_loss}). We show an example in \fref{fig:abl1-1}. For this case, some regions do not have a good match to the reference (\ie, the right "trunk" object). By using the \emph{\hmm{Chrominance branch}} only, we attain results with incorrect colors for trunk objects ($4th$ column). However, the \emph{Perceptual branch} is capable of addressing this problem ($8th$ column). It predicts the single and natural brown color for the trunk, since the majority of trunks in the training data are brown. Thus, the prediction of the \emph{Perceptual branch} is purely based on the dominant color of objects from the large-scale data, and independent of the reference. As we can see in the $8th$ column, it predicts the same colors even for different references.
 
To enjoy the advantages of both branches, we adopt a multi-task training strategy to train both branches simultaneously. The term $\alpha$ is used as their relative weight. The double-branch results in $5th-7th$ columns of \fref{fig:abl1-1} explicitly indicate that our network learns to adaptively fuse the predictions of both branches: selecting and propagating the reference color at well-matched regions, but generalizing to the natural color learnt from large-scale data for mismatched or unrelated regions. The relative weight $\alpha$ tunes the preference towards each branch. Evaluated on the ImageNet validation data, we set $\alpha=0.005$ as the default in our experiments.

\begin{figure}
\centering
\footnotesize
\setlength{\tabcolsep}{0.003\linewidth}
\begin{tabular}{cccc}
    \includegraphics[width=0.24\linewidth]
{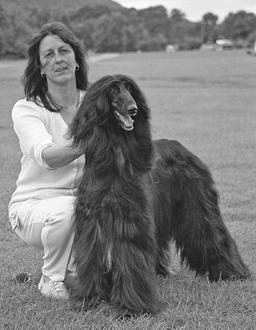}&
 \includegraphics[width=0.24\linewidth]
{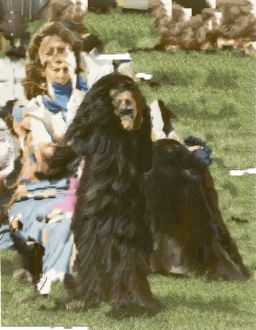}&
    \includegraphics[width=0.24\linewidth]
    {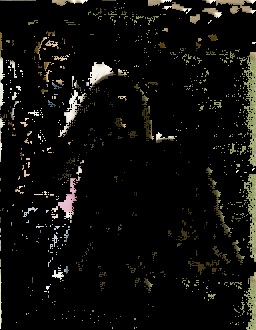}&
     \includegraphics[width=0.24\linewidth]
    {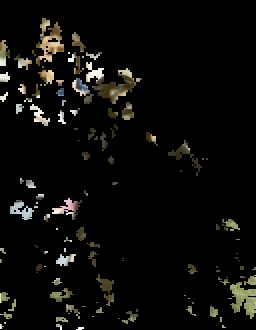}\\
      Target& Aligned reference& Samples& Samples \\
       & & (threshold) & (cross-check)\\
    \includegraphics[width=0.24\linewidth]
{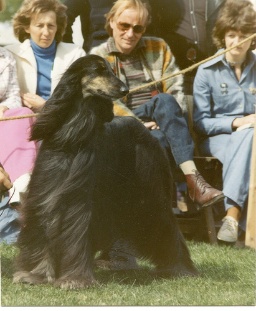}&
	 \includegraphics[width=0.24\linewidth]
{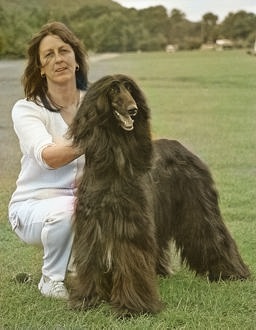}&
  \includegraphics[width=0.24\linewidth]
{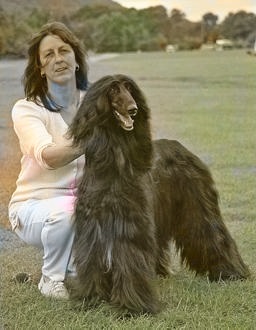} &
     \includegraphics[width=0.24\linewidth]
{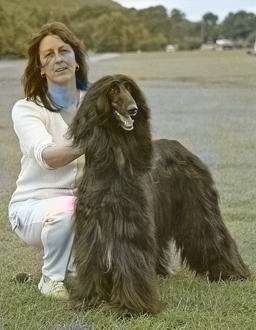}\\
 Reference& Our result& Zhang et al.~\shortcite{zhang2017real}& Zhang et al.~\shortcite{zhang2017real}\\
       & & (threshold) & (cross-check)\\
\end{tabular}
  \vspace*{-.1in}
 \caption{Comparison of our end-to-end network with the alternative of selecting color samples with manual thresholds or cross-check matching, and then colorizing with Zhang et al.~\protect\shortcite{zhang2017real}. \revision{Input images: ImageNet dataset.}}
\label{fig:abl2}
\vspace{-2em}
\end{figure}

\begin{figure*}
\footnotesize
\setlength{\tabcolsep}{0.003\linewidth}
\begin{tabular}{cccccc}
    \includegraphics[height=0.105\textwidth]
{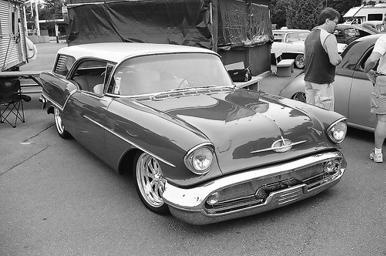}&
    \includegraphics[height=0.105\textwidth]{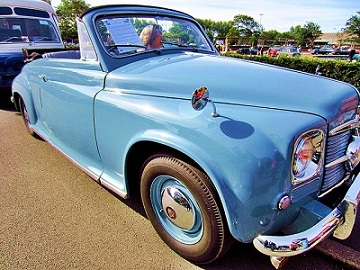}&
    \includegraphics[height=0.105\textwidth]
{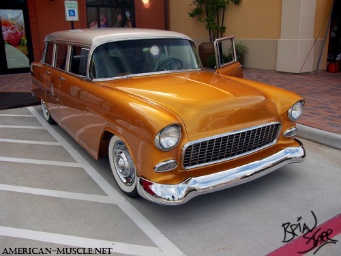}&
    \includegraphics[height=0.105\textwidth]
{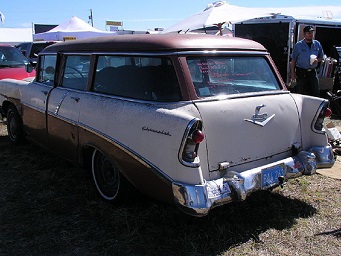}&
    \includegraphics[height=0.105\textwidth]
{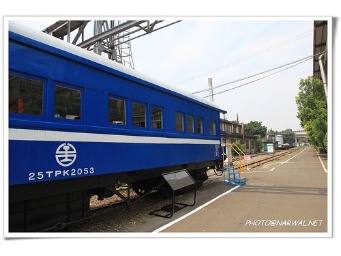}&
    \includegraphics[height=0.105\textwidth]{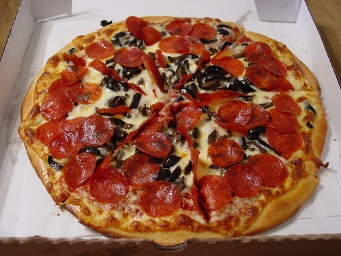}
 \\
 \includegraphics[height=0.105\textwidth]
{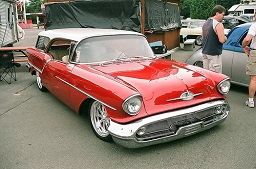}&
    \includegraphics[height=0.105\textwidth]
{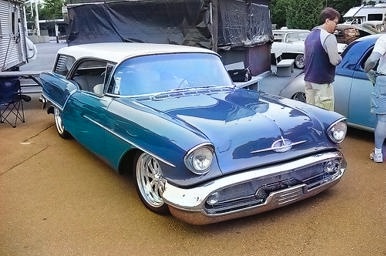}&
    \includegraphics[height=0.105\textwidth]
{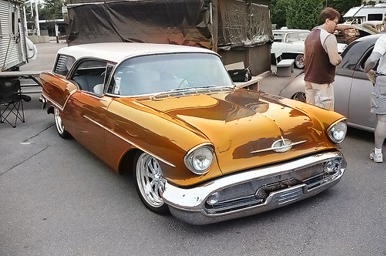}&
\includegraphics[height=0.105\textwidth]
{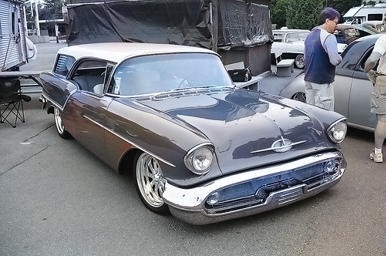}&
\includegraphics[height=0.105\textwidth]
{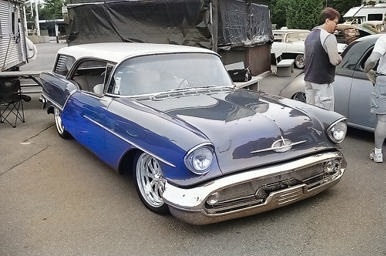}&
\includegraphics[height=0.105\textwidth]
{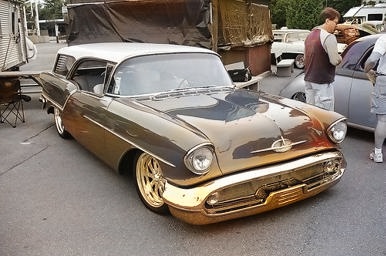}
\\
\includegraphics[height=0.118\textwidth]
{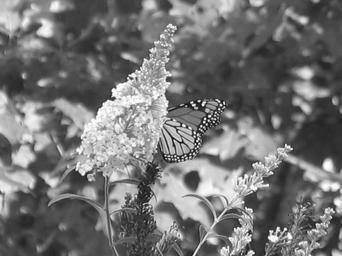}&
    \includegraphics[height=0.118\textwidth]{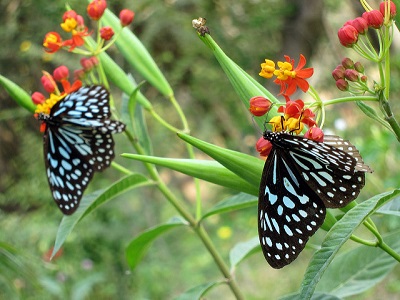}&
    \includegraphics[height=0.118\textwidth]
{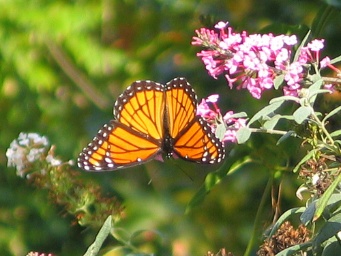}&
    \includegraphics[height=0.118\textwidth]
{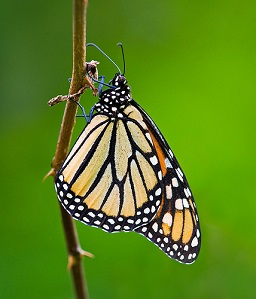}&
    \includegraphics[height=0.118\textwidth]
{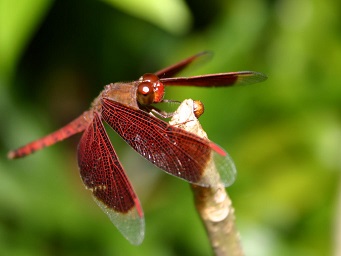}&
    \includegraphics[height=0.118\textwidth]{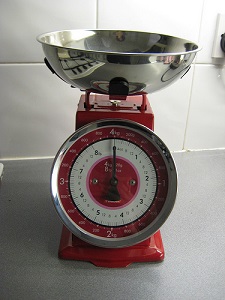}
 \\
 \includegraphics[height=0.118\textwidth]
{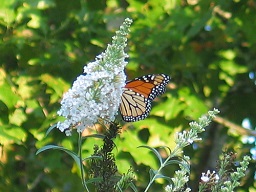}&
    \includegraphics[height=0.118\textwidth]
{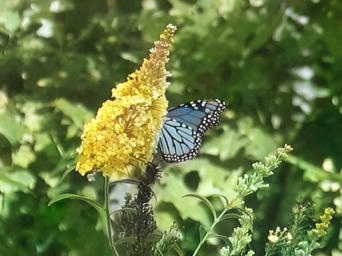}&
    \includegraphics[height=0.118\textwidth]
{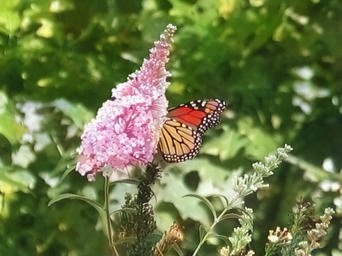}&
\includegraphics[height=0.118\textwidth]
{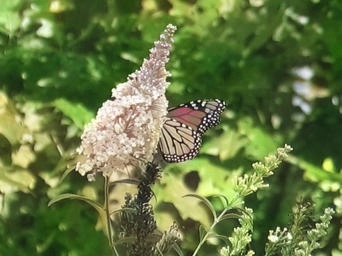}&
\includegraphics[height=0.118\textwidth]
{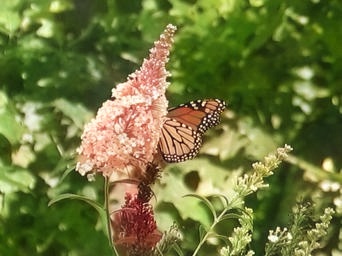}&
\includegraphics[height=0.118\textwidth]
{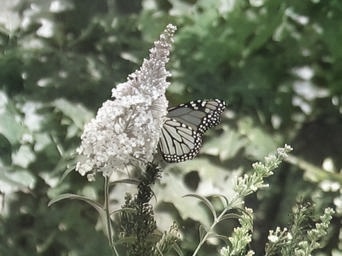}\\
 Target $\&$ Ground truth & Manually selected & Top-1 & Intra-class & Intra-category & Inter-category\\
\end{tabular}
\vspace*{-.1in}
 \caption{Our method predicts plausible colorization with different references: manually selected, automatically recommended, randomly selected in the same class of the target, randomly selected in the same category, and randomly selected out of the category. \revision{Input images: ImageNet dataset except the two manual reference photos by Andreas Mortonus/flickr and Indi Samarajiva/flickr.}}
\label{fig:abl3-1}
\vspace{-1.0em}
\end{figure*}

\begin{figure}
\footnotesize
\setlength{\tabcolsep}{0.003\linewidth}
\begin{tabular}{ccccc}
\includegraphics[height=0.2\linewidth]
{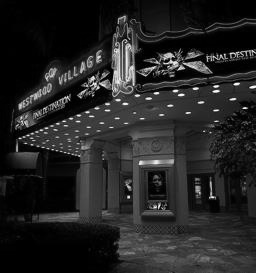}&
    \includegraphics[height=0.2\linewidth]
{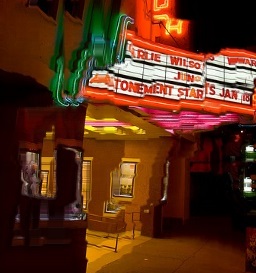}& \includegraphics[height=0.2\linewidth]
{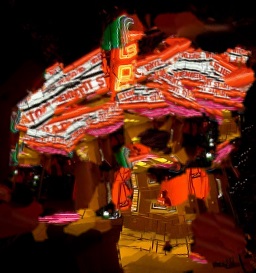}&
    \includegraphics[height=0.2\linewidth]
{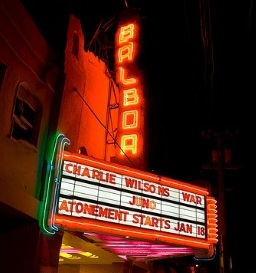}&
    \includegraphics[height=0.2\linewidth]
{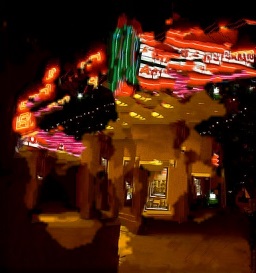} \\
  \includegraphics[height=0.2\linewidth]
{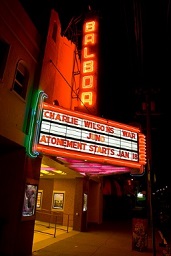}&
   \includegraphics[height=0.2\linewidth]
{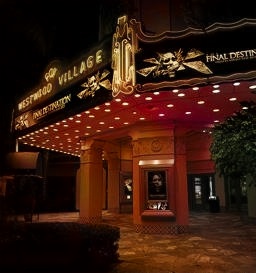}& \includegraphics[height=0.2\linewidth]
{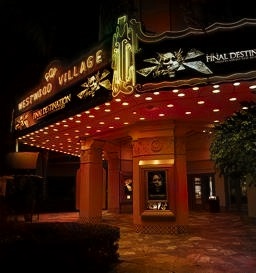}&
    \includegraphics[height=0.2\linewidth]
{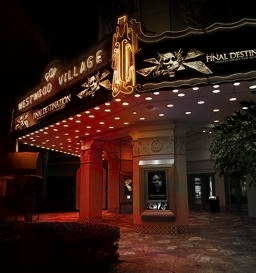}&
\includegraphics[height=0.2\linewidth]
{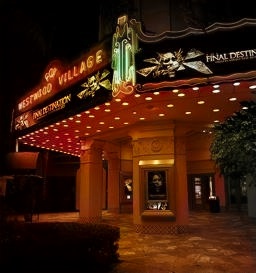}\\
Target $\&$ & SIFTFlow & DaisyFlow & DeepFlow & Deep Analogy\\
Reference &  & &  & 
\end{tabular}
 \vspace*{-.1in}
\caption{Our method works with different dense matching algorithms. \hmm{The first row shows the target and the aligned references by different matching algorithms: SIFTFlow (\protect\cite{liu2011sift}), DaisyFlow (\protect\cite{tola2010daisy}), DeepFlow (\protect\cite{weinzaepfel2013deepflow}), and Deep Image Analogy (\protect\cite{liao2017visual}). The second row shows the reference and final colorized results using different aligned references.} \revision{Input images: ImageNet dataset.}}
\label{fig:abl3-2}
\vspace{-2em}
\end{figure}

\begin{figure*}
\footnotesize
\setlength{\tabcolsep}{0.003\linewidth}
\begin{tabular}{cccccc}
     \includegraphics[width=0.155\textwidth]
{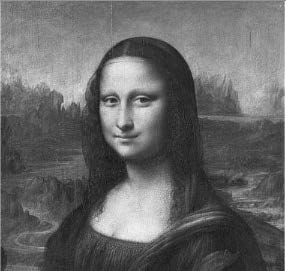}&
\includegraphics[width=0.155\textwidth]
{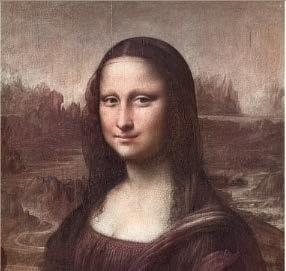}&
    \includegraphics[width=0.155\textwidth]
{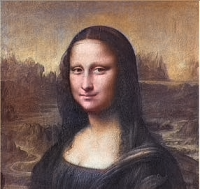}&
    \includegraphics[width=0.155\textwidth]
{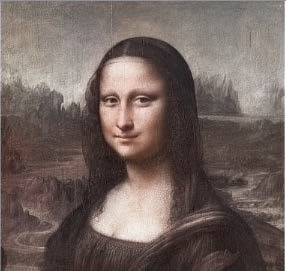}&
     \includegraphics[width=0.155\textwidth]
{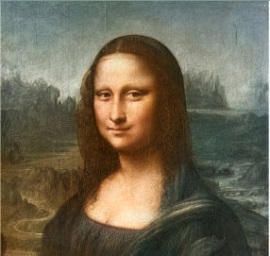} &
    \includegraphics[width=0.165\textwidth]
{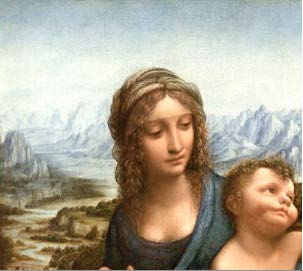}
\\
\includegraphics[width=0.155\textwidth]
{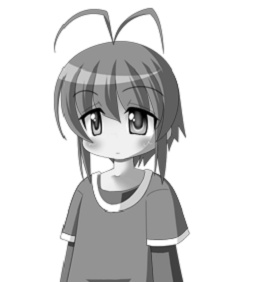}&
\includegraphics[width=0.155\textwidth]
{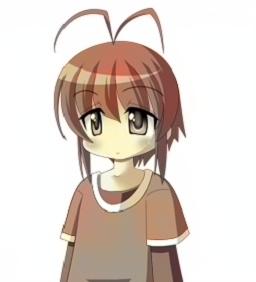}&
    \includegraphics[width=0.155\textwidth]
{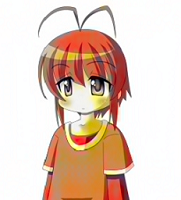}&
    \includegraphics[width=0.155\textwidth]
{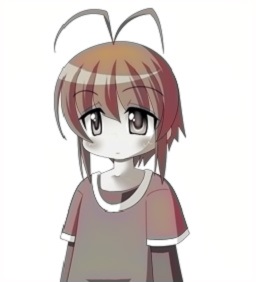}&
    \includegraphics[width=0.155\textwidth]
{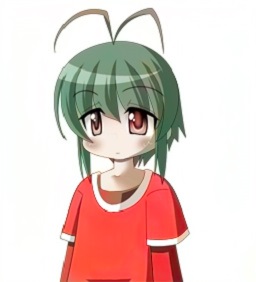} &
    \includegraphics[width=0.155\textwidth]
{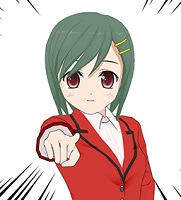}\\
 Target &Iizuka et al.~\shortcite{iizuka2016let} &Zhang et al.~\shortcite{zhang2016colorful} &Larsson et al.~\shortcite{larsson2016learning} & Ours & Reference
\end{tabular}
\vspace*{-.1in}
 \caption{Transferability comparison of colorization networks trained on ImageNet. \revision{Input images (from left to right, top to bottom): Charpiat et al.~\protect\shortcite{charpiat2008automatic}, Snow64/wikimedia and Ryo Taka/pixabay.}}
\label{fig:abl4}
\end{figure*}

\subsection{Why is end-to-end learning crucial?}

Our \emph{Colorization sub-net} learns three key components in colorization: color sample selection, color propagation, and dominant color prediction. To our knowledge, there is no other work that learns three steps simultaneously through a neural network.

An alternative is to simply sequentially process the three steps. In our study, we adopt the state-of-the-art color propagation and prediction method~\cite{zhang2017real}. Such a learning-based method significantly advances previous optimization methods~\cite{levin2004colorization}, especially when given few user points. We try two color selection strategies: 1) Threshold: select color points with the top 10$\%$ averaged bidirectional similarity score; 2) Cross-check in matching: select color points where the bidirectional mapping satisfies $\phi_{T\rightarrow R}(\phi_{R\rightarrow T})(p)=p$. Once the points are obtained, we directly feed them to the pre-trained color propagation network~\cite{zhang2017real}. We show the two predicted colorization results in $3rd$ and $4th$ columns of \fref{fig:abl2} respectively.

As we can see, the colorization does not work well and introduces many noticeable color artifacts. One possible reason is that the network~\cite{zhang2017real} is not trained on the type of input samples, but rather on user-guided points instead. Therefore, such a sequential learning would always result in a sub-optimal solution. 

Moreover, the study also shows the difficulty in determining hand-crafted rules for point selections, as mentioned in \Sref{sec:whatlearn}. It is hard to eliminate all improper color samples through heuristics. The pre-trained network will also propagate wrong samples, thus causing such artifacts. On the contrary, our end-to-end learning approach avoids these pitfalls by jointly learning selection, propagation and prediction, resulting in a single network that directly optimizes for the quality of the final colorization.

\subsection{Robustness}

A significant advantage of our network is the robustness to reference selection when compared with traditional exemplar-based colorization. It can provide plausible colors whether the reference is related or unrelated to the target. \fref{fig:abl3-1} shows how well our method works on varying references with different levels of similarity to the target image. As we can see, the colorization result is naturally more faithful to the reference when the reference is more similar to the target in their semantic content. In other situations, the result will be degenerated to a conservative colorization. This is due to the \emph{Perceptual branch}, which predicts the dominant colors from large-scale data. This behavior is similar to the existing learning-based approaches (\eg,~\cite{iizuka2016let,larsson2016learning,zhang2016colorful}).

In addition, our network is also robust to different types of dense matching algorithms, as shown in \fref{fig:abl3-2}. Note that our network is only trained using Deep Image Analogy~\cite{liao2017visual} as the default matching approach, and the network is tested with various matching algorithms. We can also observe that the result is more faithful to the reference color at well-aligned regions; while the result is degenerated to the dominant colors at misaligned regions. \hmm{Note that better alignment can improve the results of objects which can find semantic correspondences in the reference, but cannot help the colorization of objects which do not exist in the reference.}

\subsection{Transferability}

Previous learning-based methods are data-driven and thus only able to colorize images that share common properties with those in the training set. Since their networks are trained on natural images, like the ImageNet dataset, they would fail to provide satisfactory colors for unseen images, for example, human-created images (\eg, paintings or cartoons). Their results may degrade to no colorization ($1st$, $3rd$ columns in \fref{fig:abl4}) or introduce notable color artifacts ($2nd$ column). By contrast, our method benefits from the reference and successfully works in both cases. Although our network does not see such types of images in training, with the \emph{\hmm{Chrominance branch}} it learns to predict colors based on correlations of image pairs. The learnt ability is common to unseen objects.

\section{Comparison and Results}
In this section, we first report our performance and user study results. Then we qualitatively and quantitatively compare our method to previous techniques, including learning-based, exemplar-based, and interactive-based methods. Finally, we validate our method on legacy grayscale images and videos.

\subsection{Performance}
Our core algorithm is developed in CUDA. All of our experiments are conducted on a PC with an Intel E5 2.6GHz CPU and an NVIDIA Titan XP GPU. The total runtime for a $256 \times 256$ image is approximately 0.166s, including 0.016s for reference recommendation, 0.1s for similarity measurement and 0.05s for colorization.

\begin{table}[t]
\footnotesize
\caption{Colorization results compared with learning-based methods on $10,000$ images from the ImageNet validation set. The second and third columns are the Top-5 and Top-1 classification accuracies after colorization using the VGG19-BN and VGG16 network. The last column is the PSNR between the colorized result and the ground truth.}
\vspace*{.025in}
 \begin{tabular}{lccccc}
  \toprule
    & VGG19-BN/ & VGG19-BN/ &\\
     & VGG16 & VGG16 &\\
    & Top-5 Class & Top-1 Class &  \\
     & Acc($\%$) & Acc($\%$) & PSNR(dB) \\
   \midrule
   Ground truth (color) & 90.35/89.99 & 71.12 /71.25 & NA \\
   Ground truth (gray) & 84.2/81.35 & 61.5/57.39 & 23.\hmm{28} \\
 Iizuka et al.~\shortcite{iizuka2016let} & 85.53/84.12 & 63.42/61.61 & 24.\hmm{92} \\
 Zhang et al.~\shortcite{zhang2016colorful} & 84.28/83.12 & 60.97/60.25 & 22.\hmm{43} \\
 Larsson et al.~\shortcite{larsson2016learning} & 85.42/83.93 & 63.56/61.36 & \textbf{25.\hmm{50}} \\
 Ours & \textbf{85.94/84.79} & \textbf{65.1/63.73} & 22.\hmm{92} \\
  \bottomrule
 \end{tabular}
\label{tbl:psnr}
 \end{table}

\subsection{Comparison with Exemplar-based methods}
To compare with existing exemplar-based methods~\cite{welsh2002transferring,ironi2005colorization,bugeau2014variational,gupta2012image}, we run our algorithm on 35 pairs collected from their papers. \fref{fig:exm} shows several representatives \revision{and the complete set can be found in the supplemental material}. To provide a fair comparison, we directly borrow their results from their publications or run their publicly available code. 

In these examples, the content and object layouts of the reference are very similar to the target (\ie, no irrelevant objects or great intensity disparities). This is a strict requirement of existing exemplar-based methods, whose colorization relies solely on low-level features and is not learned from large-scale data. On the contrary, our algorithm is more general and has no such restrictive requirement. Even on these very related image pairs, our method shows better visual quality than previous techniques. The success comes from the sophisticated mechanism of color sample selection and propagation that are jointly learned from data rather than through heuristics.

\begin{figure*}
	\footnotesize
	\setlength{\tabcolsep}{0.003\linewidth}
	\scalebox{1.1}{
		\begin{tabular}{ccccccc}
			\includegraphics[height=0.112\linewidth]
			{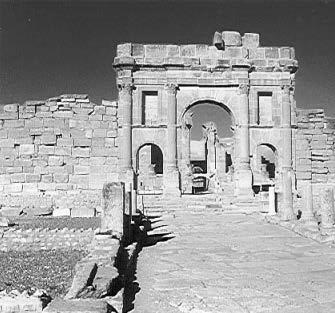}&
			\includegraphics[height=0.112\textwidth]
			{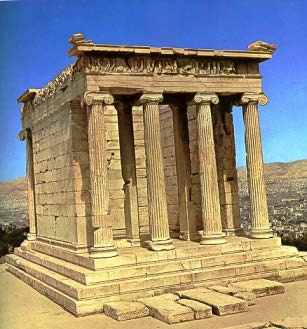}&
			\includegraphics[height=0.112\textwidth]
			{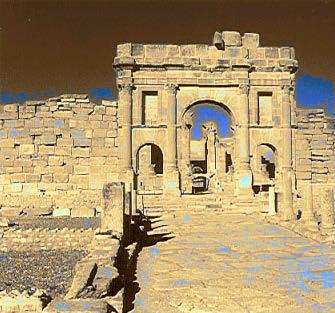}&
			\includegraphics[height=0.112\textwidth]
			{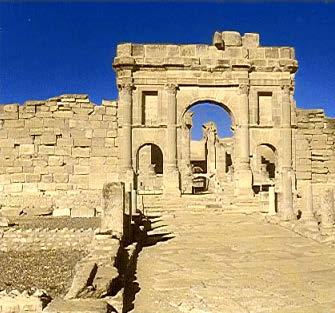}&
			\includegraphics[height=0.112\textwidth]
			{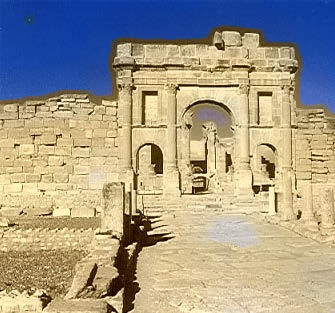} &
			\includegraphics[height=0.112\textwidth]
			{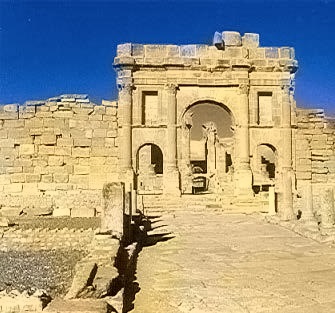}  &
			\includegraphics[height=0.112\textwidth]
			{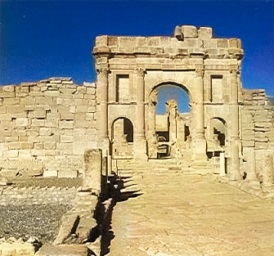}
			\\
			\includegraphics[height=0.096\textwidth]
			{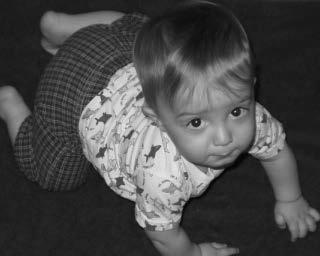}&
			\includegraphics[height=0.096\textwidth]
			{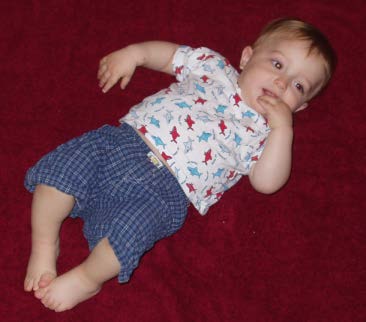}&
			\includegraphics[height=0.096\textwidth]
			{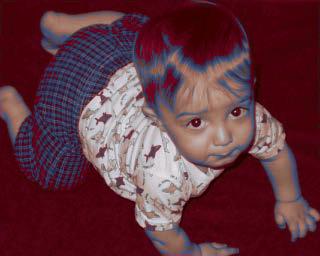}&
			\includegraphics[height=0.096\textwidth]
			{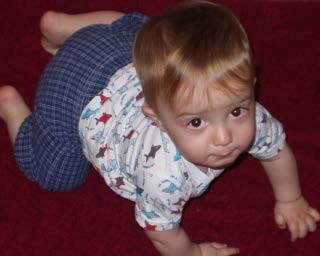}&
			\includegraphics[height=0.096\textwidth]
			{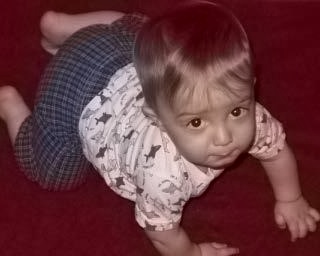} &
			\includegraphics[height=0.096\textwidth]
			{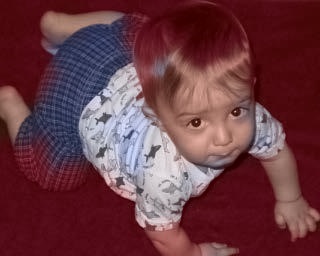} &
			
			\includegraphics[height=0.096\textwidth]
			{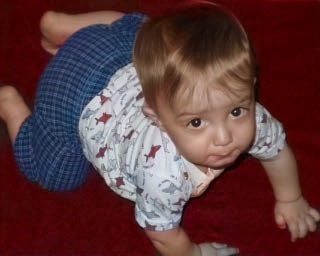} 
			\\
			\includegraphics[width=0.12\textwidth]
			{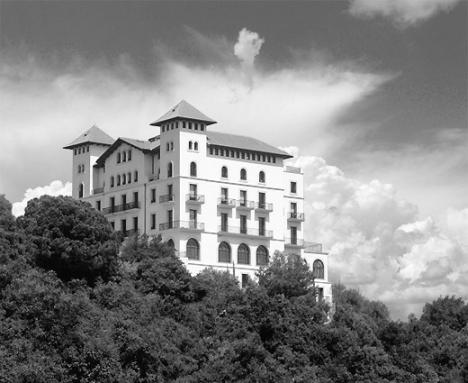}&
			\includegraphics[width=0.12\textwidth]
			{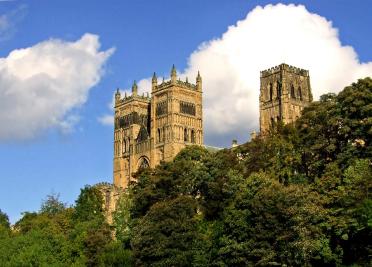}&
			\includegraphics[width=0.12\textwidth]
			{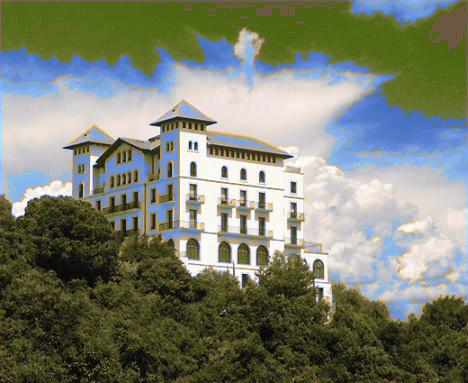}&
			\includegraphics[width=0.12\textwidth]
			{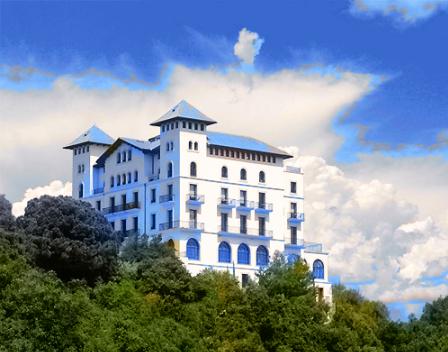}&
			\includegraphics[width=0.12\textwidth]
			{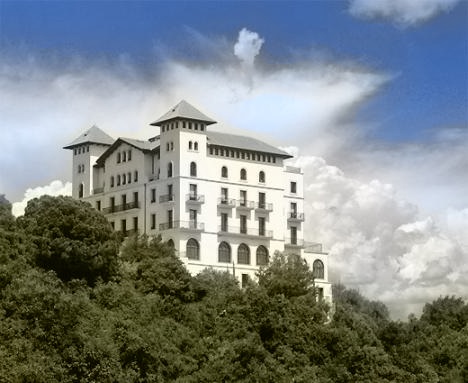} &
			\includegraphics[width=0.12\textwidth]
			{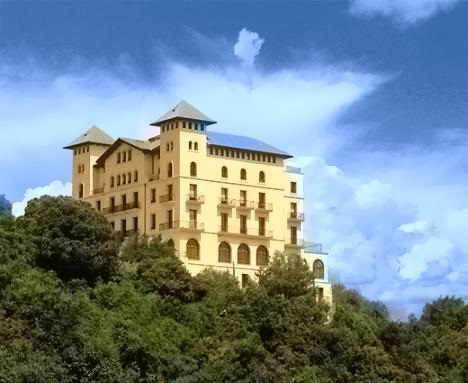} &
			\includegraphics[width=0.12\textwidth]
			{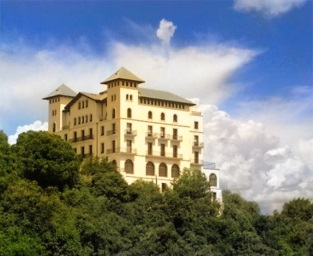}
			\\
			\includegraphics[height=0.123\textwidth]
			{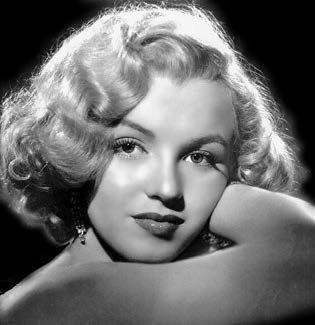}&
			\includegraphics[height=0.123\textwidth]
			{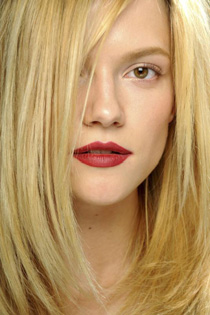}&
			\includegraphics[height=0.123\textwidth]
			{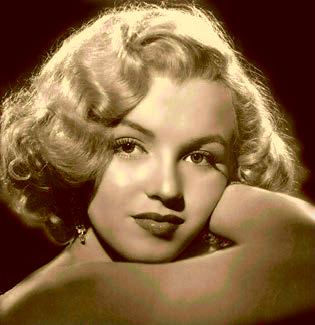}&
			\includegraphics[height=0.123\textwidth]
			{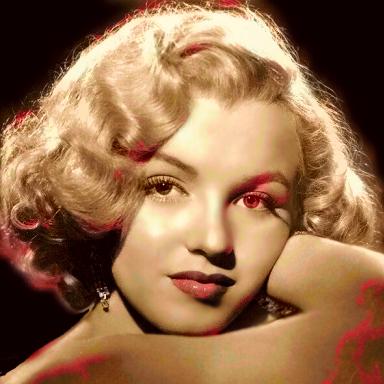}&
			\includegraphics[height=0.123\textwidth]
			{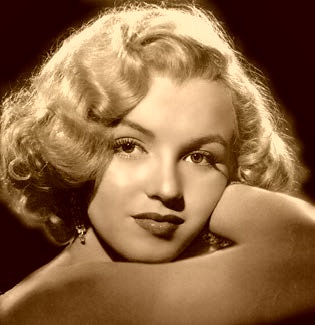} &
			\includegraphics[height=0.123\textwidth]
			{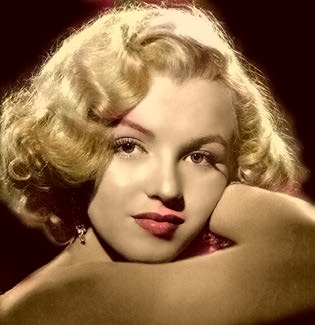} &
			\includegraphics[height=0.123\textwidth]
			{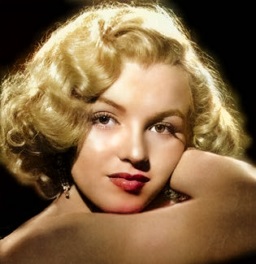}
			\\
			\includegraphics[height=0.075\textwidth]
			{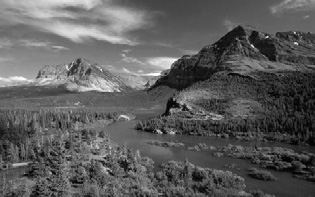}&
			\includegraphics[height=0.075\textwidth]
			{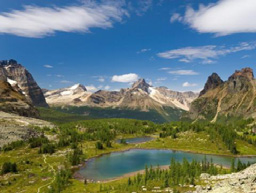}&
			\includegraphics[height=0.075\textwidth]
			{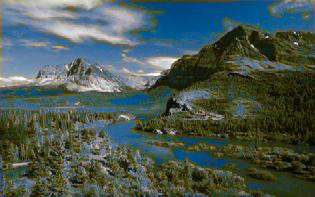}&
			\includegraphics[height=0.075\textwidth]
			{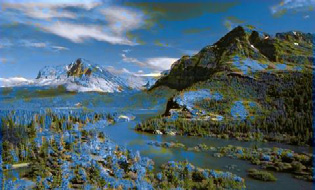}&
			\includegraphics[height=0.075\textwidth]
			{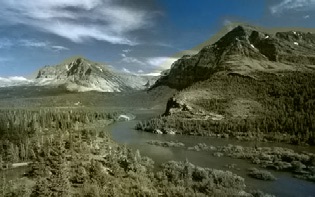} &
			\includegraphics[height=0.075\textwidth]
			{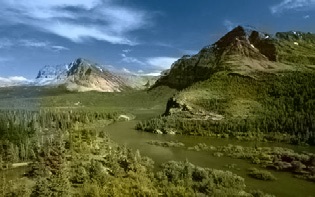} &
			\includegraphics[height=0.075\textwidth]
			{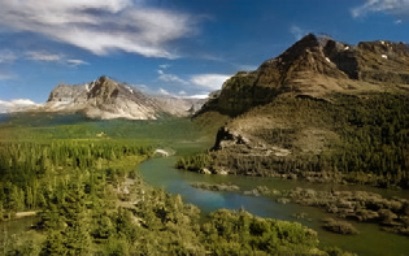}
			\\
			Target & Reference &Welsh et al.~\shortcite{welsh2002transferring} &Ironi et al.~\shortcite{ironi2005colorization} &Bugeau et al.~\shortcite{bugeau2012patch} &Gupta et al.~\shortcite{gupta2012image} & Ours\\
	\end{tabular}}
	\vspace*{-.1in}
	\caption{Comparison results with example-based methods. \revision{Input images: Ironi et al.~\protect\shortcite{ironi2005colorization} and Gupta et al.~\protect\shortcite{gupta2012image}.}}
	\label{fig:exm}
\end{figure*}

\begin{figure*}
	\footnotesize
	\setlength{\tabcolsep}{0.003\linewidth}
	\scalebox{0.96}
	{\begin{tabular}{ccccccc}
			\includegraphics[width=0.14\textwidth]
			{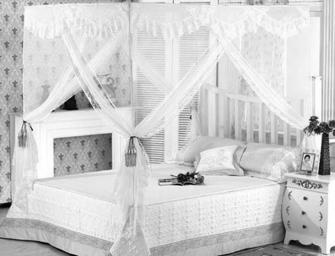}&
			\includegraphics[width=0.14\textwidth]
			{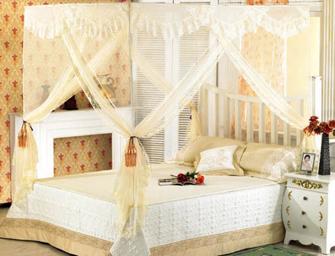}&
			\includegraphics[width=0.14\textwidth]
			{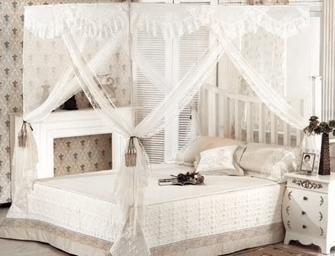}&
			\includegraphics[width=0.14\textwidth]
			{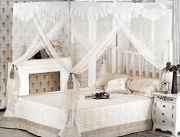}&
			\includegraphics[width=0.14\textwidth]
			{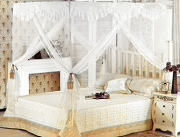}&
			\includegraphics[width=0.14\textwidth]
			{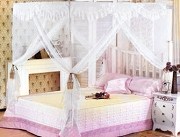} &
			\includegraphics[width=0.132\textwidth]
			{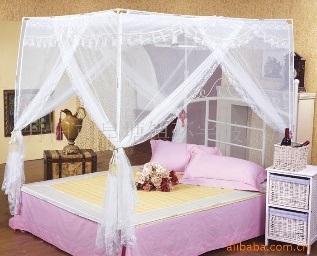} 
			\\
			\includegraphics[width=0.14\textwidth]
			{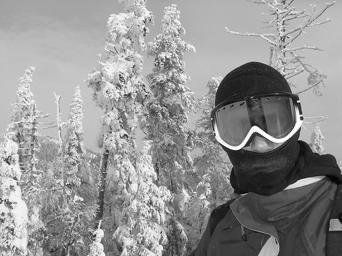}&
			\includegraphics[width=0.14\textwidth]
			{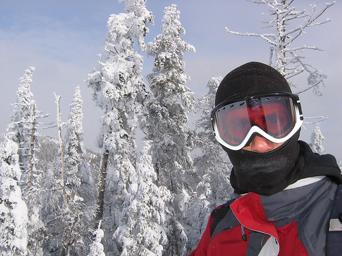}&
			\includegraphics[width=0.14\textwidth]
			{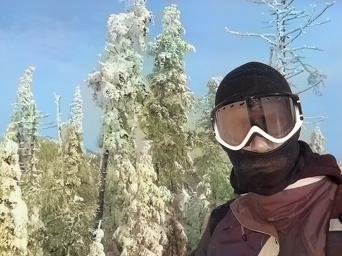}&
			\includegraphics[width=0.14\textwidth]
			{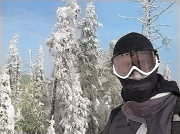}&
			\includegraphics[width=0.14\textwidth]
			{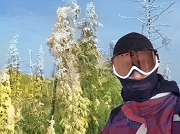}&
			\includegraphics[width=0.14\textwidth]
			{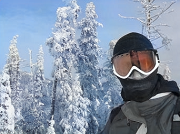} &
			\includegraphics[width=0.138\textwidth]
			{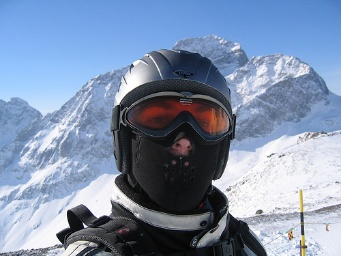}  
			\\
			\includegraphics[width=0.14\textwidth]
			{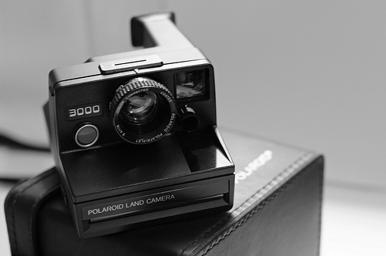}&
			\includegraphics[width=0.14\textwidth]
			{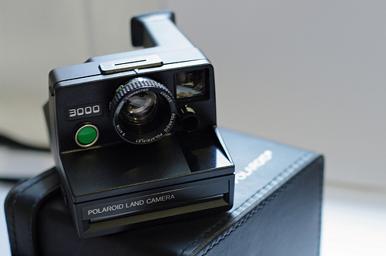}&
			\includegraphics[width=0.14\textwidth]
			{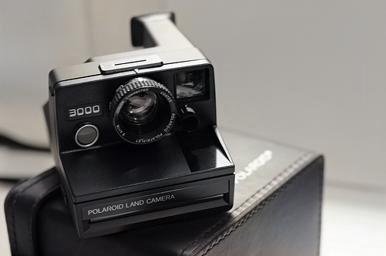}&
			\includegraphics[width=0.14\textwidth]
			{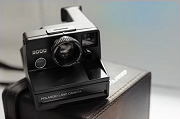}&
			\includegraphics[width=0.14\textwidth]
			{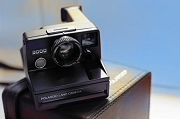}&
			\includegraphics[width=0.14\textwidth]
			{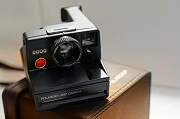} &
			\includegraphics[width=0.135\textwidth]
			{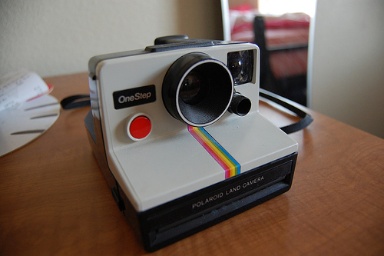} 
			\\
			\includegraphics[width=0.14\textwidth]
			{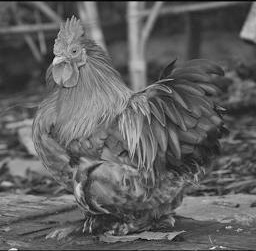}&
			\includegraphics[width=0.14\textwidth]
			{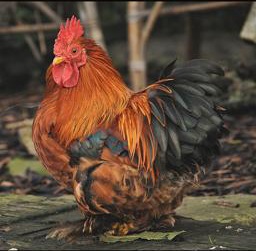}&
			\includegraphics[width=0.14\textwidth]
			{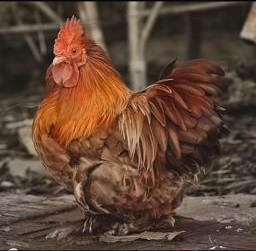}&
			\includegraphics[width=0.14\textwidth]
			{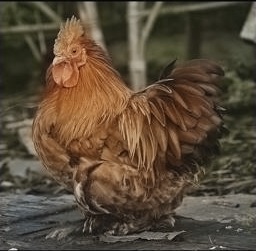}&
			\includegraphics[width=0.14\textwidth]
			{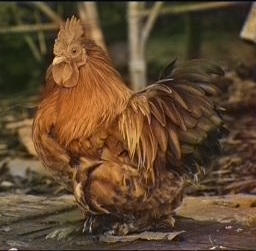}&
			\includegraphics[width=0.14\textwidth]
			{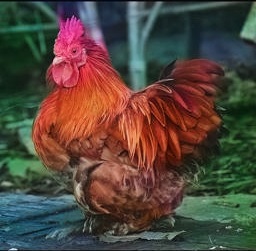} &
			\includegraphics[width=0.103\textwidth]
			{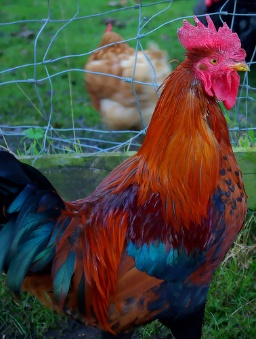} 
			\\
			\includegraphics[width=0.14\textwidth]
			{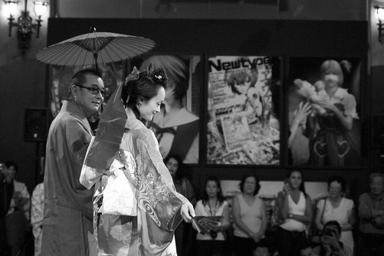}&
			\includegraphics[width=0.14\textwidth]
			{images/learning_based_comp/028_in}&
			\includegraphics[width=0.14\textwidth]
			{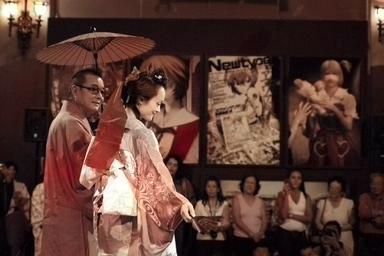}&
			\includegraphics[width=0.14\textwidth]
			{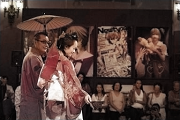}&
			\includegraphics[width=0.14\textwidth]
			{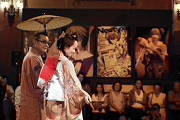}&
			\includegraphics[width=0.14\textwidth]
			{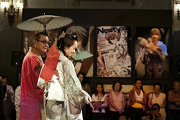} &
			\includegraphics[width=0.135\textwidth]
			{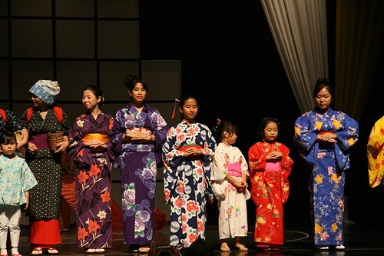} 
			\\
			\includegraphics[width=0.14\textwidth]
			{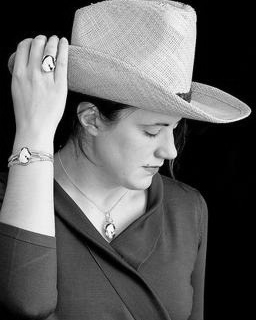}&
			\includegraphics[width=0.14\textwidth]
			{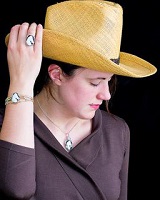}&
			\includegraphics[width=0.14\textwidth]
			{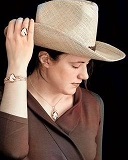}&
			\includegraphics[width=0.14\textwidth]
			{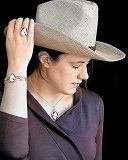}&
			\includegraphics[width=0.14\textwidth]
			{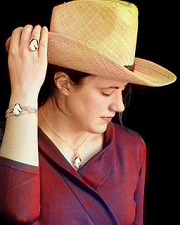}&
			\includegraphics[width=0.14\textwidth]
			{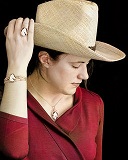} &
			\includegraphics[width=0.14\textwidth]
			{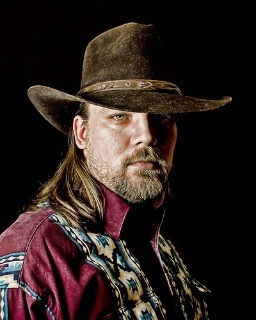} 
			\\
			\includegraphics[width=0.14\textwidth]
			{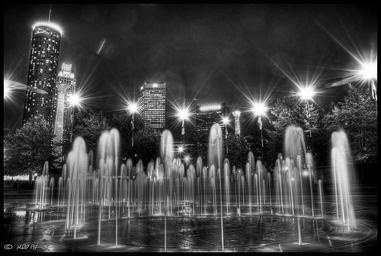}&
			\includegraphics[width=0.14\textwidth]
			{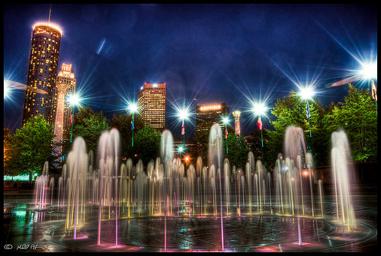}&
			\includegraphics[width=0.14\textwidth]
			{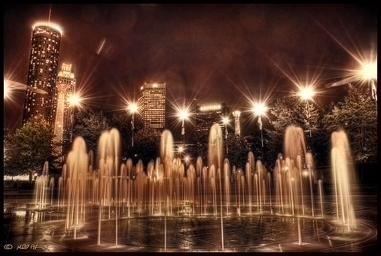}&
			\includegraphics[width=0.14\textwidth]
			{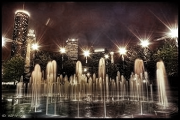}&
			\includegraphics[width=0.14\textwidth]
			{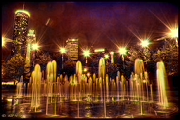}&
			\includegraphics[width=0.14\textwidth]
			{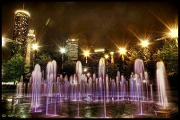} &
			\includegraphics[width=0.14\textwidth]
			{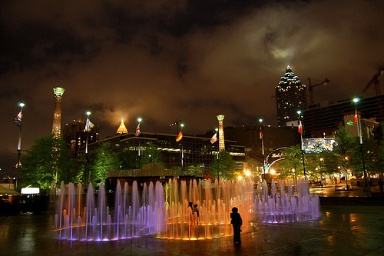} 
			\\
			Target & Ground truth &Iizuka et al.~\shortcite{iizuka2016let} & Larsson et al.~\shortcite{larsson2016learning} & Zhang et al.~\shortcite{zhang2016colorful} & Ours & Reference\\
		\end{tabular}
	}
	\vspace*{-.1in}
	\caption{Comparison results with learning-based methods. \revision{Input images: ImageNet dataset.}}
	\label{fig:learning}
\end{figure*}
\begin{figure*}
\footnotesize
\setlength{\tabcolsep}{0.001\linewidth}
\begin{tabular}{ccccccc}
\includegraphics[height=0.14\linewidth]
{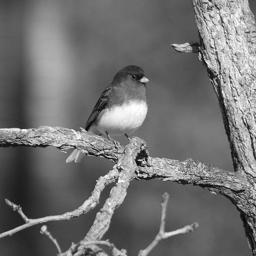}&
\includegraphics[height=0.14\linewidth]
{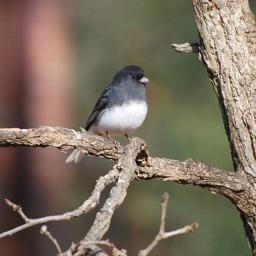}&
\includegraphics[height=0.14\linewidth]
{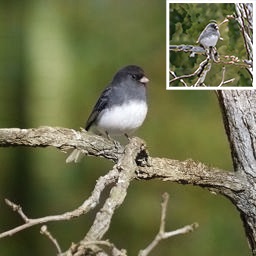}&
\includegraphics[height=0.14\linewidth]
{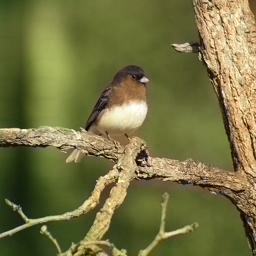}&
\includegraphics[height=0.14\linewidth]
{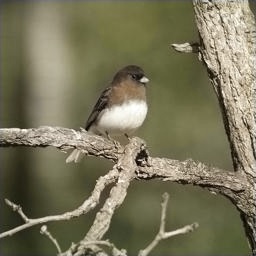}&
\includegraphics[height=0.14\linewidth]
{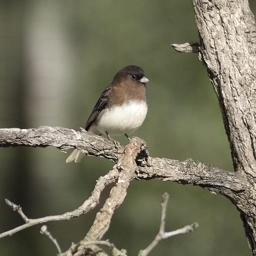}&
\includegraphics[height=0.14\linewidth]
{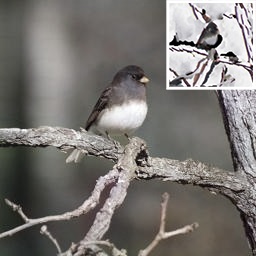}\\
Target & Ground truth & Ours (Top-1 ref) &Zhang et al.~\shortcite{zhang2017real} & Larsson et al.~\shortcite{larsson2016learning} &Iizuka et al.~\shortcite{iizuka2016let} & Ours (random ref)\\
& & ($72\%$) & ($65\%$)& ($42\%$)& ($38\%$) & ($7\%$)\\
\end{tabular}
\vspace*{-.1in}
\caption{\hmm{An example to show users preference on vibrant colorization. The numbers in brackets represent its fooling rates. Our colorized results ($3rd$ and last columns) are guided by the top-right references.} \revision{Input images: ImageNet dataset.}}
\label{fig:rank}
\end{figure*}

\subsection{Comparison with learning-based methods}
We compare our method with the-state-of-the-art learning-based colorization networks~\cite{larsson2016learning,zhang2016colorful,iizuka2016let} by evaluating on $10,000$ images in the validation set of ImageNet (same as Larsson et al.~\shortcite{larsson2016learning}). Our method is trained on a subset of the ImageNet training set, as described in \Sref{sec:col}. We tested our automatic solution by taking the Top-1 recommendation as the reference (\sref{sec:rec}). To be fair, we use author-released models trained on the ImageNet dataset as well to run their methods.

We show a quantitative comparison of colorized results in Table \ref{tbl:psnr} on two metrics: PSNR relative to the ground truth and classification accuracy. Our results have a lower PSNR score (22.9178dB) than Larsson et al.~\shortcite{larsson2016learning} and Iizuka et al.~\shortcite{iizuka2016let}, because PSNR overly penalizes a plausible but different colorization result. A correct colorization faithful to the reference may even achieve a lower PSNR than a conservative colorization, such as predicting gray for every pixel (24.9214dB). On the contrary, our method outperforms all other methods on image recognition accuracy rates when sending the colorized results into VGG19 or VGG16 pre-trained on image recognition task. It indicates that our colorized results seem to be more natural than others, which can be recognizable as well as the true color image.

A qualitative comparison for selected representative cases is shown in \fref{fig:learning}. \revision{For a full set of $200$ images randomly drawn from $1,000$ cases, please refer to our supplemental material.} From this comparison, an apparent difference is that our results are more saturated and colorful when compared to Iizuka et al.~\shortcite{iizuka2016let} and Larsson et al.~\shortcite{larsson2016learning}, with the help of sampling colorful points from the reference. Zhang et al.~\shortcite{zhang2016colorful} uses a class-rebalancing step to oversample more colorful portions of the gamut during training, but such a solution sometimes results in overly aggressive colorization and causes artifacts (\eg the blue and orange colors in the 4th row of \fref{fig:learning}). Our approach can control colorization and achieve desired colors by simply giving different references, thus our results are visually faithful to the reference colors.

\hmm{In addition to quantitative and qualitative comparisons, we use a perceptual metric to evaluate how compelling our colorization looks to a human observer. We ran a \emph{real vs. fake} two-alternative forced-choice user study on Amazon Mechanical Turk (AMT) across different learning-based methods. This is similar to the approach taken by Zhang et al.~\shortcite{zhang2016colorful}. Participants in the study were shown a series of pairs of images. Each pair consisted of a ground-truth color photo and a re-colorized version produced by either our algorithm (randomly selected reference or Top-1 recommended reference) or a baseline~\cite{iizuka2016let,larsson2016learning,zhang2016colorful}. The two images were shown side-by-side in randomized order. For every pair, participants were asked to observe the image pair for no more than 5 seconds and click on the photo they believed was the most realistic as early as possible. All images were shown with the resolution of $256$ pixels on the short edge.}

\hmm{To guarantee all algorithms can be compared by the same "turker" populations, we included results from different algorithms in one experimental session for each participant. Each session consisted of 5 practice trials (excluded from subsequent analysis), followed by 50 randomly selected test pairs (each algorithm contributed 10 pairs). During the practice trials, participants were given feedback as to whether their answers were correct. No feedback was given during the 50 test pairs. We conducted 5 different sessions to make sure every algorithm covered all image pairs. The participants were only allowed to complete at most one session. All experiment sessions were posted simultaneously and a total of 125 participants were involved in the user study (25$\pm2$ participants per session).}

\hmm{As shown in \Tref{tbl:fool}, our method with the Top-1 reference ($38.08\%$) and Zhang et al.~\shortcite{zhang2016colorful} ($35.36\%$) respectively ranked $1st$ and $2nd$ in the fooling rate. We \revision{felt} that this may be partly because participants preferred more colorful results to less saturated results as shown in \fref{fig:rank}. Zhang et al.~\shortcite{zhang2016colorful} uses a class-rebalancing step to encourage rare colors but at the expense of images which are \revision{overly-aggressively} colorized; while our method produces more vibrant colorization by utilizing correct color samples from the reference. Our method with random reference also degenerates to conservative color prediction since few reliable color samples can be used from the unrelated reference. This verifies that a good reference is important to high-quality colorization.}

\begin{figure}
\footnotesize
\setlength{\tabcolsep}{0.003\linewidth}
\includegraphics[width=0.95\linewidth]
{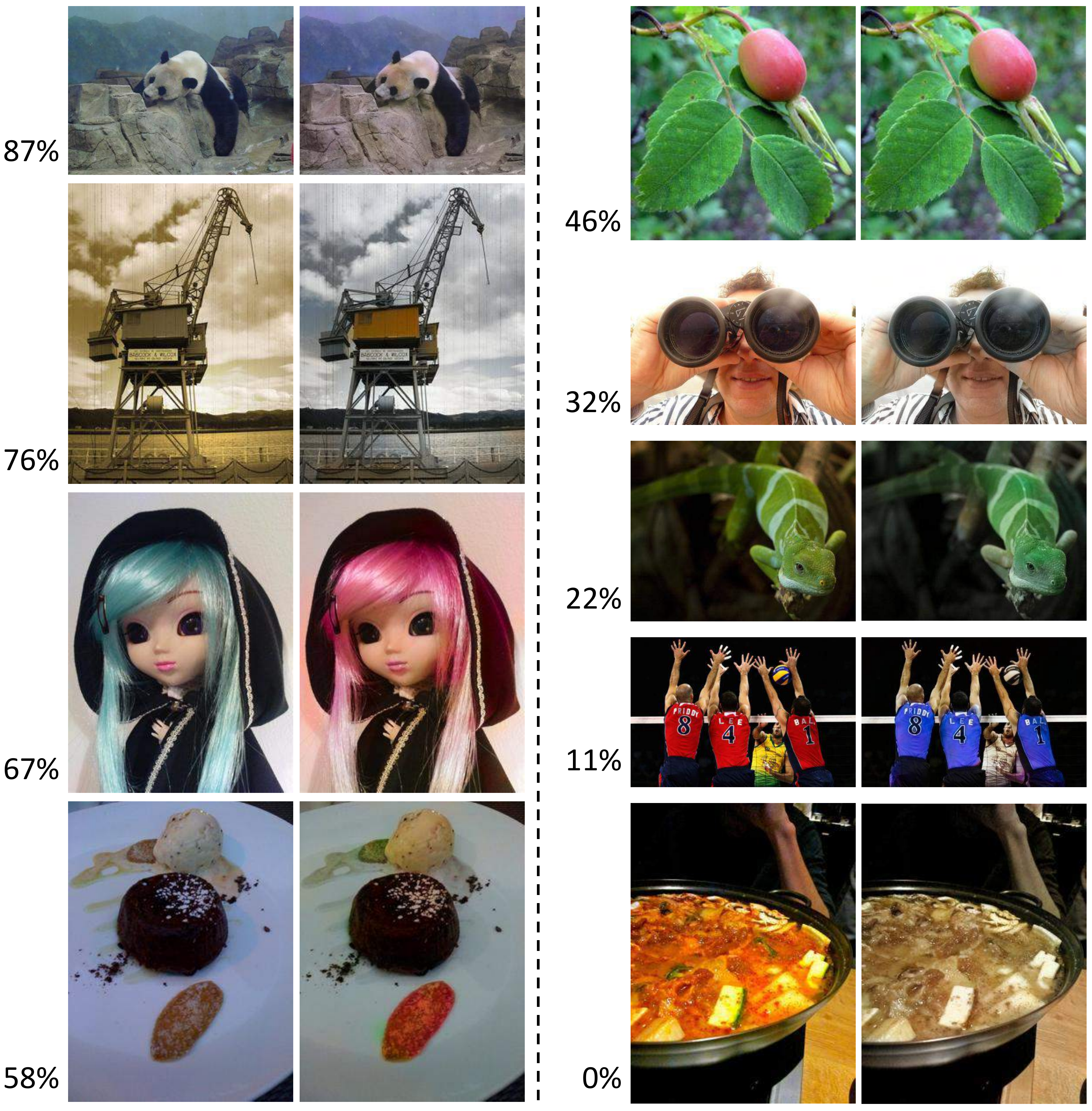}
\begin{tabular}{cccc}
\quad \enskip Ground truth \quad \quad & \quad Our result \quad \quad \quad & \quad \quad \enskip Ground truth \quad \quad & \quad Our result \quad \\ 
\end{tabular}
 \vspace*{-.1in}
 \caption{Examples from the user study. Results are generated with our method with the Top-1 reference and are sorted by how often the users chose our algorithm's colorization over the ground truth. \revision{Input images: ImageNet dataset.}}
  \label{fig:user_rank}
  \vspace{-2em}
\end{figure}

\fref{fig:user_rank} provides a better sense of the participants' competency at detecting subtle errors made by our algorithm. The percentage on the left shows how often participants think our colorized result is more realistic than the ground truth. Some issues may come from lack of colorization in some local regions (\eg, $0\%, 11\%$), or poor white balancing in the ground truth image (\eg, $22\%, 32\%$). Surprisingly, our results are considered more natural to human observers than the ground truth image in some cases (\eg $87\%, 76\%$).

\begin{table}[t]
\caption{\hmm{Amazon Mechanical Turk real v.s. fake fooling rate. We compared our method using an automatically recommended reference or a random intra-class reference with other learning-based methods.
 Note that the best expectation of fooling rate should be around $50\%$, which occurs when the user cannot distinguish real from fake images and is forced to choose between two equally believable images. Input images: ImageNet dataset.}}
  \vspace*{.025in}
 \begin{tabular}{lc}
  \toprule
    Method \qquad \qquad & \qquad \qquad  Fooling Rate ($\%$) \\
  \midrule
 Iizuka et al.~\shortcite{iizuka2016let} \qquad \qquad & \qquad \qquad  24.56 $\pm$ 1.76\\
  Larsson et al.~\shortcite{larsson2016learning} \qquad \qquad & \qquad \qquad 24.64 $\pm$ 1.71\\
  Zhang et al.~\shortcite{zhang2016colorful} \qquad \qquad & \qquad \qquad 35.36 $\pm$ 1.52\\
   Ours with random reference \qquad \qquad & \qquad \qquad 21.92 $\pm$ 1.56\\
   Ours with Top-1 reference \qquad \qquad & \qquad \qquad \textbf{38.08} $\pm$ \textbf{1.72}\\   
  \bottomrule
 \end{tabular}
\label{tbl:fool}
\end{table}
 
\subsection{Comparison with interactive-based methods}
\begin{figure}[h]
	\footnotesize
	\setlength{\tabcolsep}{0.003\linewidth}
	\begin{tabular}{cccc}
		\includegraphics[width=0.222\linewidth]
		{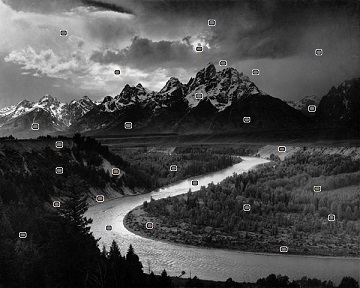}&
		\includegraphics[width=0.222\linewidth]
		{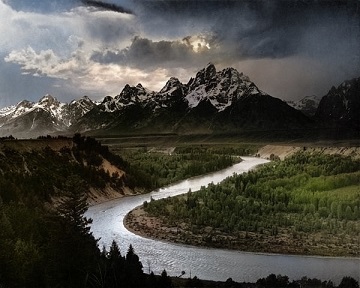}&
		\includegraphics[width=0.222\linewidth]
		{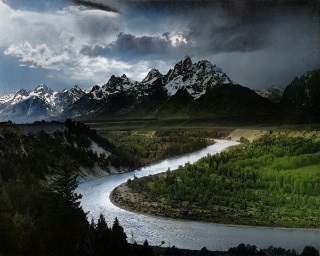}&
		\includegraphics[width=0.267\linewidth]
		{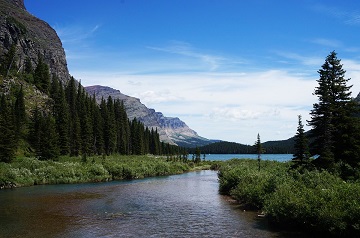}
	\end{tabular}
	\begin{tabular}{cccc}
		\includegraphics[height=0.24\linewidth]
		{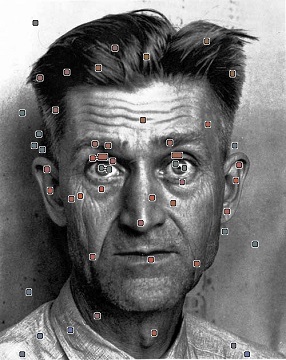}&
		\includegraphics[height=0.24\linewidth]
		{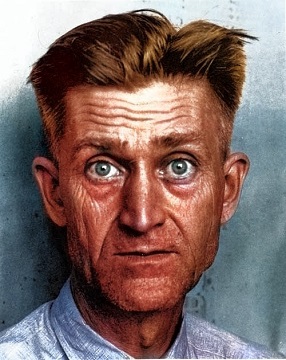}&
		\includegraphics[height=0.24\linewidth]
		{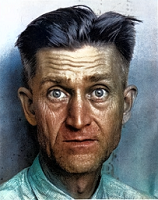}&
		\includegraphics[height=0.24\linewidth]
		{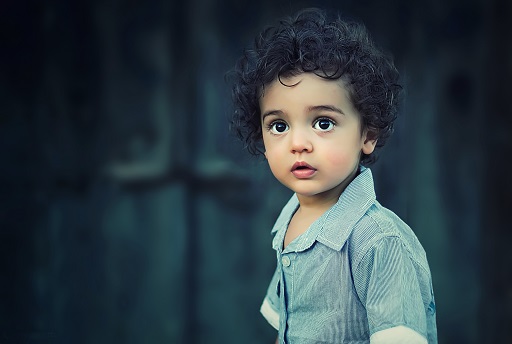}\\
		Target &Zhang et al.~\shortcite{zhang2017real} & Ours & Reference\\
	\end{tabular}
	\vspace*{-.1in}
	\caption{Comparison results with the interactive-based method. The points overlaid on the target are manually given and used in Zhang et al.~\protect\shortcite{zhang2016colorful}, while the reference in the last column is manually selected and used by our approach. \revision{Input images (from left to right, top to bottom): Ansel Adams/wikipedia, Carina Chen/pixabay, Dorothea Lange and Bess Hamiti/pixabay.}}
	\label{fig:inter}
	\vspace{-1em}
\end{figure}

We compare our hybrid method with a different hybrid solution~\cite{zhang2017real} which combines user-guided scribbles (\ie, points) and deep learning. As shown in \fref{fig:inter}, by giving a proper reference selected by the user, our method can achieve comparable quality to theirs with dozens of user-given color points. Thus, our method proposes a simple way to control the appearance of colorization generated with the help of deep neural networks.

\begin{figure}[h]
	\centering
	\footnotesize
	\setlength{\tabcolsep}{0.003\linewidth}
	\begin{tabular}{cccc}
		\includegraphics[width=0.24\linewidth]
		{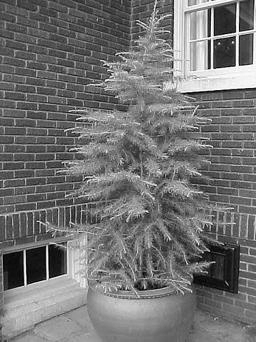}&
		\includegraphics[width=0.24\linewidth]
		{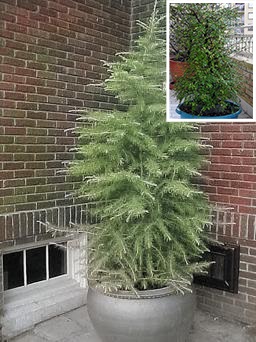}&
		\includegraphics[width=0.24\linewidth]
		{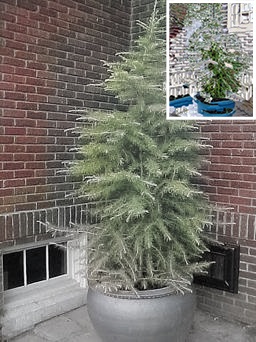}&
		\includegraphics[width=0.24\linewidth]
		{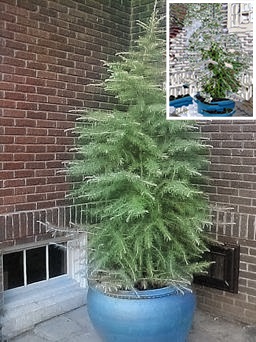}\\
		Target &Zhang et al.~\shortcite{zhang2017real} &Zhang et al.~\shortcite{zhang2017real} & Ours\\
		& (Top-1 ref) & (Top-1 aligned ref) & (Top-1 ref)\\
		\includegraphics[width=0.24\linewidth]
		{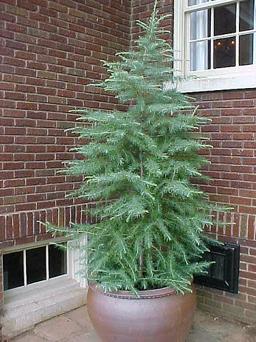}&
		\includegraphics[width=0.24\linewidth]
		{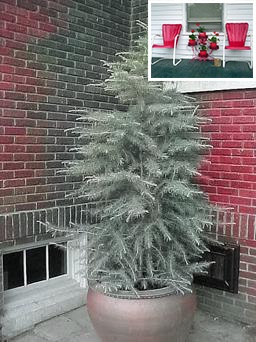}&
		\includegraphics[width=0.24\linewidth]
		{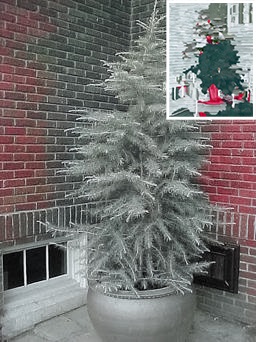} &
		\includegraphics[width=0.24\linewidth]
		{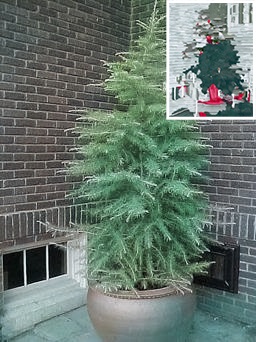}\\
		Ground truth &Zhang et al.~\shortcite{zhang2017real} &Zhang et al.~\shortcite{zhang2017real} & Ours\\
		&(Random ref) & (Random aligned ref) & (Random ref)\\
	\end{tabular}
	\vspace*{-.1in}
	\caption{\hmm{Comparison to Zhang et al.~\protect\shortcite{zhang2016colorful} using global histogram hints from references overlaid on the the top-right corner. The histogram used in Zhang et al.~\protect\shortcite{zhang2016colorful} is either from the reference ($2nd$ column) or from its aligned version generated by Liao et al.~\protect\shortcite{liao2017visual} ($3rd$ column).} \revision{Input images: ImageNet dataset.}}
	\label{fig:global}
	\vspace{-1em}
\end{figure}

\hmm{Zhang et al.~\shortcite{zhang2017real} also present a variant of their method which uses a global color histogram of a reference image as input
to control colorization results. In \fref{fig:global}, we show a comparison with results by Zhang et al.~\shortcite{zhang2017real} using the global color histogram either from the reference image ($2nd$ column) or the aligned reference ($3rd$ column). Their method provides a global control to alter color distribution and average saturation but fails to achieve locally variant colorization effects. Our method can preserve semantic correspondence and locally map the reference color to the target (e.g., the plant colorized green and the flowerpot colorized in blue).}

\subsection{Colorization of legacy photographs and movies}

\begin{figure*}[h]
	\includegraphics[width=0.95\linewidth]
	{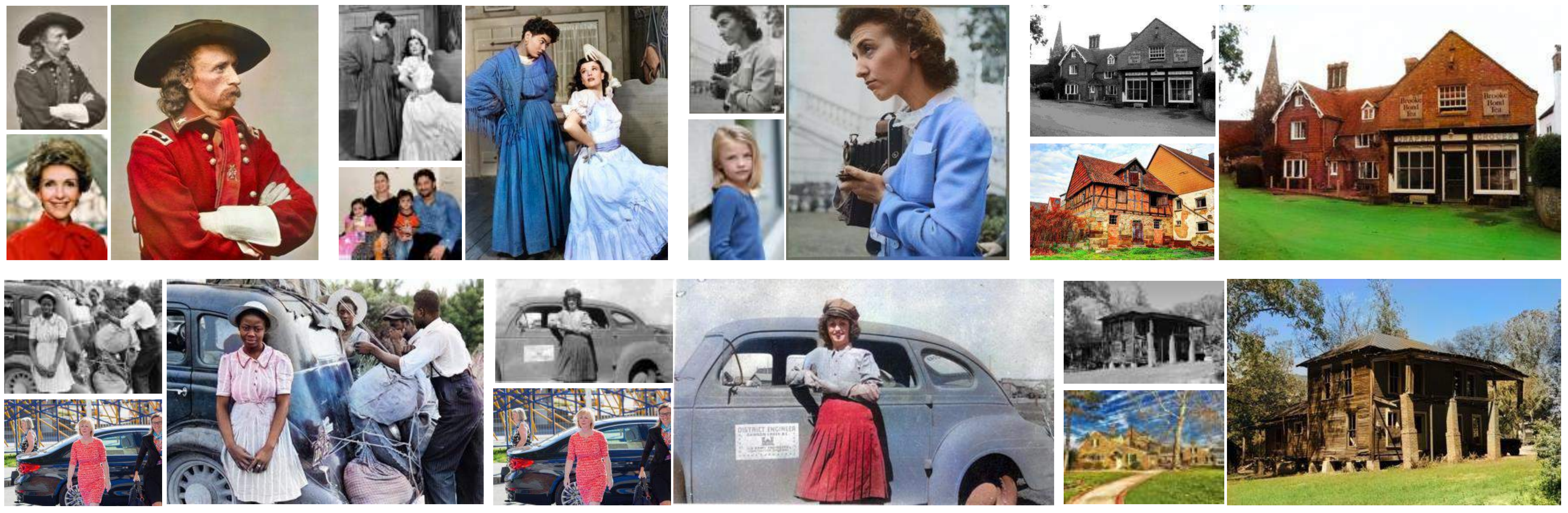} 
	\vspace*{-.1in}
	\caption{Colorization of legacy pictures. In each set, the target grayscale photo is the upper-left, the reference is the lower-left and our result lies on the right. \revision{Input images (from left to right, top to bottom, target to reference): George L. Andrews/wikipedia, Official White House Photographer/wikimedia, Vandamm/wikimedia, Anonymous/wikimedia, Esther Bubley/wikimedia, Anonymous/wikimedia, Nick Macneill/geograph, Bernd/pixabay, Oberholster Venita/pixabay, EU2017EE Estonian Presidency/wikimedia, Audrey Coey/flickr, EU2017EE Estonian Presidency/wikimedia, Patrick Feller/wikimedia and Anonymous/pixabay.}}
	\label{fig:legacy}
	\vspace{-0em}
\end{figure*}

\begin{figure*}[h]
	\footnotesize
	\setlength{\tabcolsep}{0.003\linewidth}
	\begin{tabular}{ccccc}
		&
		\includegraphics[width=0.194\textwidth]
		{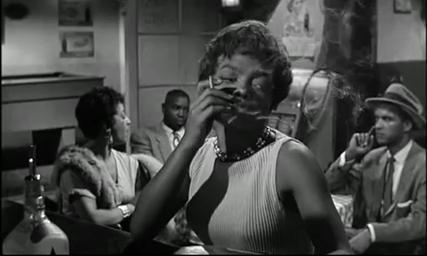}&
		\includegraphics[width=0.194\textwidth]
		{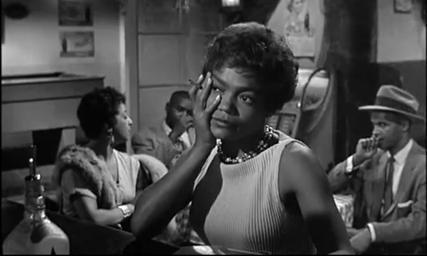}&
		\includegraphics[width=0.194\textwidth]
		{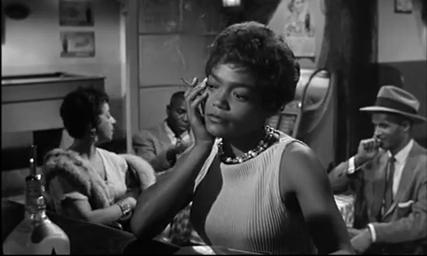}&
		\includegraphics[width=0.194\textwidth]
		{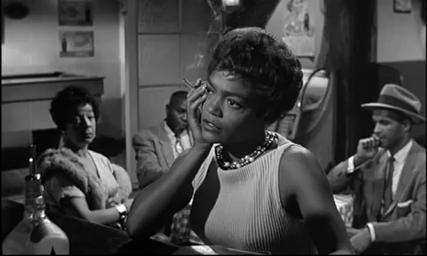}\\
		\includegraphics[width=0.1585\textwidth]
		{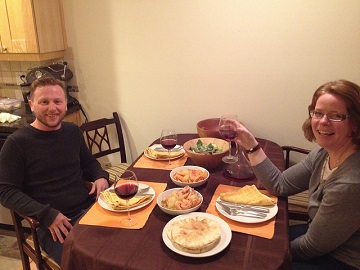}&
		\includegraphics[width=0.194\textwidth]
		{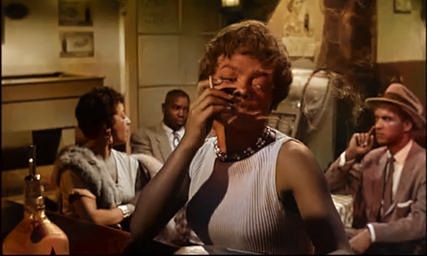}&
		\includegraphics[width=0.194\textwidth]
		{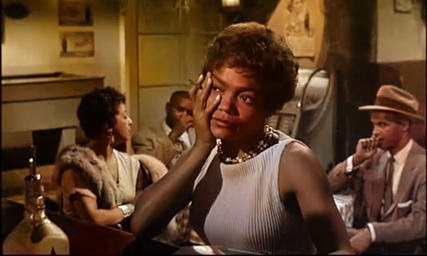}&
		\includegraphics[width=0.194\textwidth]
		{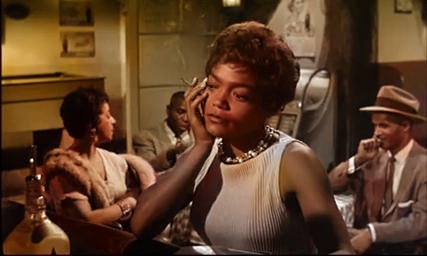}&
		\includegraphics[width=0.194\textwidth]
		{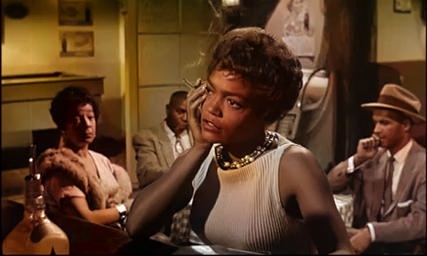}
	\end{tabular}
	\vspace*{-.1in}
	\caption{Extending our method to video colorization. All black and white frames (top row) are independently colorized with the same reference (leftmost column of bottom row) to generate colorized results (\hmm{right} 4 columns of bottom row). \revision{The input clip is from the film Anna Lucasta (public domain) and the reference photo is by Heather Harvey/flickr.}}
	\label{fig:video}
	\vspace{-1em}
\end{figure*}

Our system was trained on "synthetic" grayscale images by removing the chrominance channels from color images. We tested our system on legacy grayscale images, and show some selected results in \fref{fig:legacy}. Moreover, our method can be extended to colorize legacy movies by independently colorizing each frame and then temporally smoothing the colorized results with the method of Bonneel et al.~\shortcite{bonneel2015blind}. Some selected frames of a movie example are shown in \fref{fig:video}. \revision{Please refer to our supplemental material for a video demo.}

\section{Limitations and Conclusions}
We have presented a novel colorization approach that employs a deep learning architecture and a reference color image. Our approach is a general solution for exemplar-based colorization since it yields plausible results even in cases where the target image does not have clear correspondences in the reference. In such cases, it is still capable of producing plausible and natural colors for the target image. Unlike most deep-learning colorization frameworks, our approach allows us to control colorized results. Furthermore, with the reference recommendation algorithm, the system also provides the user with an automatic tool for re-coloring black-and-white photographs and movies.
\begin{figure}[h]
	\footnotesize
	\setlength{\tabcolsep}{0.003\linewidth}
	\begin{tabular}{ccccc}
		\includegraphics[width=0.19\linewidth]
		{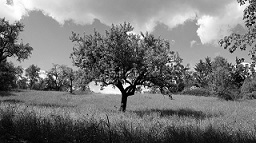}&
		\includegraphics[width=0.182\linewidth]
		{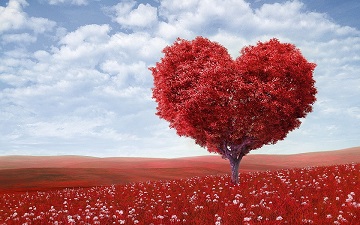}&
		\includegraphics[width=0.19\linewidth]
		{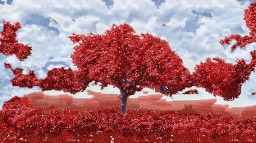}&
		\includegraphics[width=0.19\linewidth]
		{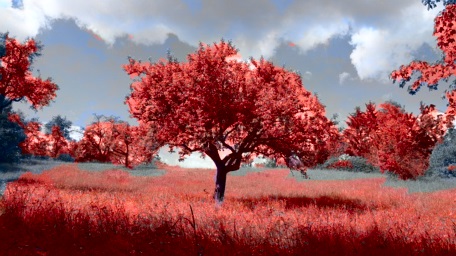}&
		\includegraphics[width=0.19\linewidth]
		{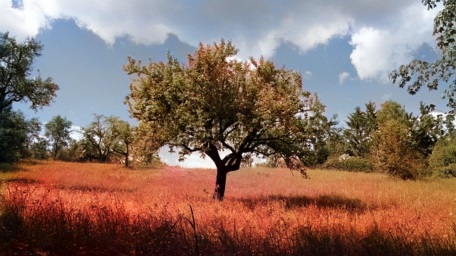}
		\\
		\includegraphics[width=0.19\linewidth]
		{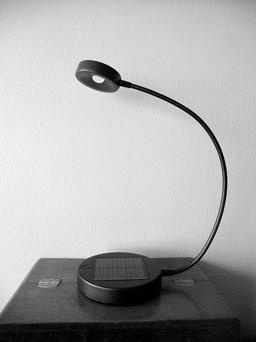}&
		\includegraphics[width=0.19\linewidth]
		{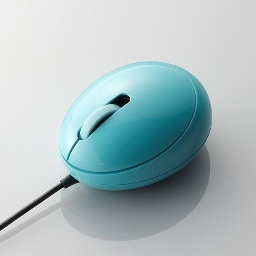}&
		\includegraphics[width=0.19\linewidth]
		{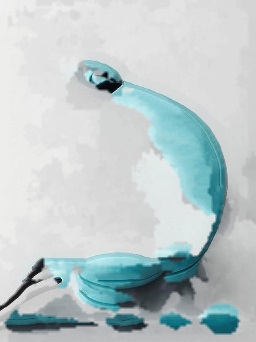}&
		\includegraphics[width=0.19\linewidth]
		{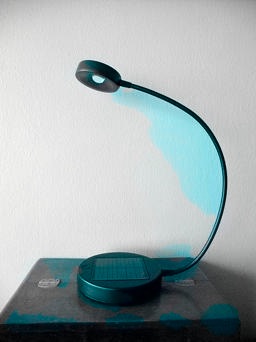}&
		\includegraphics[width=0.19\linewidth]
		{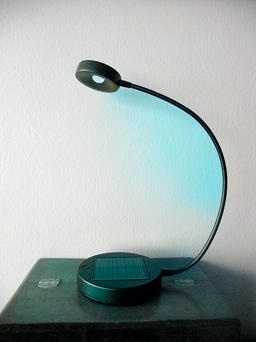}
		\\
		\includegraphics[width=0.19\linewidth]
		{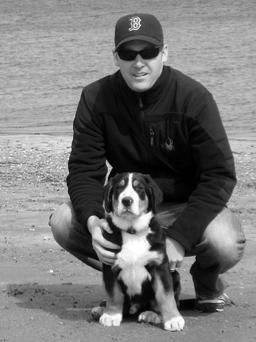}&
		\includegraphics[width=0.19\linewidth]
		{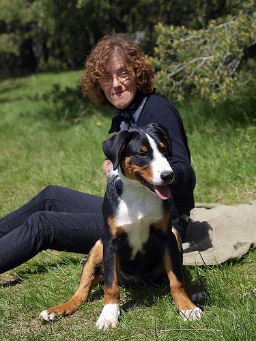}&
		\includegraphics[width=0.19\linewidth]
		{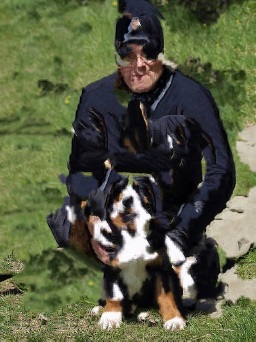}&
		\includegraphics[width=0.19\linewidth]
		{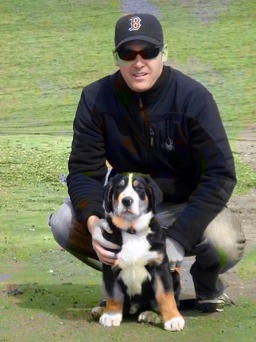}&
		\includegraphics[width=0.19\linewidth]
		{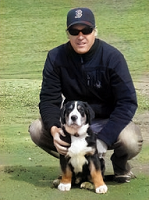}
		\\
		\includegraphics[width=0.19\linewidth]
		{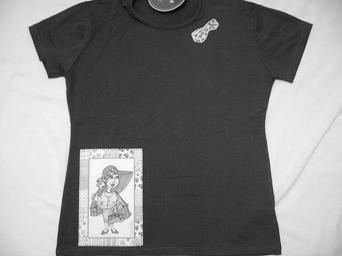}&
		\includegraphics[width=0.168\linewidth]
		{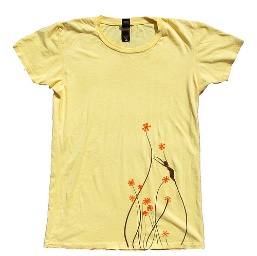}&
		\includegraphics[width=0.19\linewidth]
		{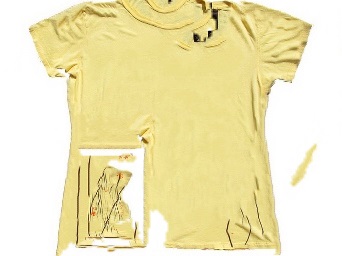}&
		\includegraphics[width=0.19\linewidth]
		{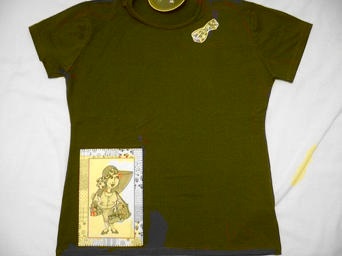}&
		\includegraphics[width=0.19\linewidth]
		{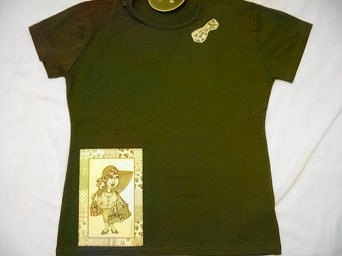}
		\\
		Target & Reference & Aligned & Combination & Result\\
		$T_L$ & $R$ & reference $R'$ & $ T_L \bigoplus R'_{ab}$&P\\
	\end{tabular}
	\vspace*{-.15in}
	\caption{Limitations of our work. Top row: our network cannot colorize objects with unusual or artistic colors constrained by the perceptual loss. \hmm{Second} row: the perceptual loss does not sufficiently penalize incorrect reference colors on regions with less semantic importance, e.g. a smooth background. \hmm{Third row: the classification network fails to distinguish regions with similar local textures, e.g. sand and grass.} \hmm{Forth} row: the result is visually less faithful to the reference if their luminance gaps are too large. \revision{Input images: ImageNet dataset except the images on the first row by Anonymous/pxhere and Anonymouse/pixabay.}}
	\label{fig:limit}
\end{figure}
Our approach also suffers from some limitations that can be addressed in future work. First, our network cannot colorize objects with unusual or artistic colors, since it is constrained by the learning from the proposed \emph{Perceptual branch}, as shown in the top row of \fref{fig:limit}. 

Second, the perceptual loss based on the classification network (VGG) cannot penalize incorrect colors in regions with less semantic importance, such as the wall in the \hmm{second} row of \fref{fig:limit}, \hmm{or fails to distinguish less semantic regions with similar local texture, such as the similar sand and grass textures in the third row of \fref{fig:limit}.} In addition, our result is less faithful to the reference when there are dramatic luminance disparities between images, as shown in the bottom row of \fref{fig:limit}. To mitigate this limitation, our reference recommendation algorithm enforces luminance similarity in the local ranking. \hmm{Occasionally}, our method fails to predict colors for some \hmm{local} regions, as shown in \fref{fig:user_rank}. It would be worthwhile to explore how to better balance the two branches of our network.

\bibliographystyle{acmsiggraph}
\nocite{*}
\bibliography{aaatemplate}
\end{document}